\documentclass{article}

\usepackage{microtype}
\usepackage{graphicx}
\usepackage{subfigure}
\usepackage{booktabs} % for professional tables
\usepackage[cmex10]{amsmath}
\usepackage{amsthm}
\usepackage{xcolor}
\usepackage{tikz} 
\usetikzlibrary{arrows.meta, calc, decorations.markings, math, arrows}

\usepackage{subfigure}

\newtheorem{theorem}{Theorem}

\newtheorem{lemma}[theorem]{Lemma}

\newtheorem{definition}{Definition}

\usepackage{amssymb}

\newcommand{\field}[1]{\mathbb{#1}}
\newcommand{\suchthat}{\;\ifnum\currentgrouptype=16 \middle\fi|\;}

\newcommand{\cB}{{\cal B}}

\newcommand{\cD}{{\cal D}}

\newcommand{\cF}{{\cal F}}
\newcommand{\cG}{{\cal G}}
\newcommand{\cH}{{\cal H}}

\newcommand{\cK}{{\cal K}}

\newcommand{\cM}{{\cal M}}

\newcommand{\cO}{{\cal O}}

\DeclareMathAlphabet{\pazocal}{OMS}{zplm}{m}{n}
\newcommand{\bv}{{\bf v}}

\newcommand{\fR}{\field{R}} % real numbers
\newcommand{\rh}{\hat{r}}

\newcommand{\mathify}[1]{\ifmmode{#1}\else\mbox{$#1$}\fi}

\newcommand{\E}[1]{\mathify{\mathbb{E}\left[#1\right]}}

\newcommand{\EX}[1]{\mathify{\mathbb{E}_{X}\left[#1\right]}}
\newcommand{\EZ}[1]{\mathify{\mathbb{E}_{Z}\left[#1\right]}}

\newcommand{\Vol}{\mathrm{Vol}}

\newcommand{\bzero}[1]{0}

\usepackage{hyperref}

\usepackage[accepted]{icml2021}

\icmltitlerunning{WGAN with an Infinitely Wide Generator Has No Spurious Stationary Points}

\begin{document}

\twocolumn[
\icmltitle{WGAN with an Infinitely Wide Generator Has No Spurious Stationary Points}




\begin{icmlauthorlist}
\icmlauthor{Albert No}{HU}
\icmlauthor{TaeHo Yoon}{ed}
\icmlauthor{Sehyun Kwon}{ed}
\icmlauthor{Ernest K. Ryu}{ed}
\end{icmlauthorlist}

\icmlaffiliation{HU}{Department of Electronic and Electrical Engineering, Hongik University, Seoul, Korea}
\icmlaffiliation{ed}{Department of Mathematical Sciences, Seoul National University, Seoul, Korea}

% \icmlcorrespondingauthor{Cieua Vvvvv}{c.vvvvv@googol.com}
\icmlcorrespondingauthor{Ernest K. Ryu}{ernestryu@snu.ac.kr}

\icmlkeywords{Deep learning theory, Generative adversarial networks, Infinitely wide networks, Universal approximation theory, Random feature learning}

\vskip 0.3in
]



\printAffiliationsAndNotice{}  % leave blank if no need to mention equal contribution

\begin{abstract}
Generative adversarial networks (GAN) are a widely used class of deep generative models, but their minimax training dynamics are not understood very well. In this work, we show that GANs with a 2-layer infinite-width generator and a 2-layer finite-width discriminator trained with stochastic gradient ascent-descent have no spurious stationary points. We then show that when the width of the generator is finite but wide, there are no spurious stationary points within a ball whose radius becomes arbitrarily large (to cover the entire parameter space) as the width goes to infinity.
\end{abstract}

\section{Introduction}
Generative adversarial networks (GAN) \citep{goodfellow_gan_2014}, which learn a generative model mimicking the data distribution, have found a broad range of applications in machine learning.
While supervised learning setups solve minimization problems in training, GANs solve \emph{minimax} optimization problems.
However, the minimax training dynamics of GANs are poorly understood.
Empirically, training is tricky to tune, as reported in \citep[Section~1]{gan_training_mescheder2018} and \citep[Section~5.1]{goodfellow_tutorial}.
Theoretically, prior analyses of minimax training have established few guarantees.

In the supervised learning setup, the limit in which the deep neural networks' width is infinite has been utilized to analyze the training dynamics.
Another line of work establishes guarantees showing that no spurious local minima exist, and such results suggest (but do not formally guarantee) that training converges to the global minimum despite the non-convexity.

In this work, we study a Wasserstein GAN (WGAN) \citep{bottou2017_wass_gan,wass_gan_training2017} with an infinitely wide generator trained with stochastic gradient ascent-descent.
Specifically, we show that a WGAN with a 2-layer generator and a 2-layer discriminator both with random features and sigmoidal%
\footnote{We say an activation function is \emph{sigmoidal} if it satisfies assumptions \hyperlink{cond:AG}{(AG)} and \hyperlink{cond:AD}{(AD)}, which we later state. The standard sigmoid and $\tanh$ activations functions are sigmoidal.}
activation functions and with the width of the generator (but not the discriminator) being large or infinite has no spurious stationary points%
\footnote{A stationary point is spurious if it is not a global minimum.}
when trained with stochastic gradient ascent-descent. 
The theoretical analysis utilizes ideas from universal approximation theory and random feature learning.

\subsection{Prior work}
The classical universal approximation theorem establishes that a 2-layer neural network with a sigmoidal activation function can approximate any continuous function when the hidden layer is sufficiently wide \citep{Cybenko1989}. This universality result was extended to broader classes of activation functions \citep{HORNIK1991251,LESHNO1993}, and quantitative bounds on the width of such approximations were established \citep{SAF_1980-1981____A5_0,barron1993universal,jones1992}.
Random feature learning \citep{rahimi2007random,4797607,kitchensink2008} combines these ingredients into the following implementable algorithm: generate the hidden layer weights randomly and optimize the weights of the output layers while keeping the hidden layer weights fixed.
% Our work uses results from the universal approximation theory and ideas from random feature learning.

In recent years, there has been intense interest in the analysis of infinitely wide neural networks,  primarily in the realm of supervised learning.
In the ``lazy training regime'', infinitely wide neural networks behave as Gaussian processes at initialization \citep{neal1996,lee2018deep}
and are essentially linear in the parameters, but not the inputs, during training. The limiting linear network can be characterized with the neural tangent kernel (NTK) \citep{NTK2018,du2018gradient,NEURIPS2018_54fe976b}.
% Jacot, Gabriel, and Hongler

In a different ``mean-field regime'', the training dynamics of infinitely wide 2-layer neural networks are characterized with a Wasserstein gradient flow.
This idea was concurrently developed by several groups \citep{chizat2018,MeiE7665,rotskoff2018,Rotskoff2019,Sirignano2020,Sirignano20202}.
Specifically relevant to GANs, this mean-field machinery was applied to study the dynamics of finding mixed Nash equilibria of zero-sum games \citep{mean-field-gan2020}.
Finally, \citet{Geiger_2020} provides a unification of the NTK and mean-field limits.

Another line of analysis in supervised learning establishes that no spurious local minima, non-global local minima, exist. 
% Such results imply that training finds a global minimum despite non-convexity.
The first results of this type were established for the matrix and tensor decomposition setups \citep{ge2016,ge2017,wu2018,Sanjabi2019}.
Later, these analyses were extended to neural networks through the notion of no spurious ``basins''
\citep{nguyen2018on,pmlr-v80-liang18a,NEURIPS2018_a0128693,Li2021,sun2020_review,9194023} and ``mode connectivity'' \citep{NEURIPS2018_be3087e7,NEURIPS2019_46a4378f,pmlr-v119-shevchenko20a}.

Prior works have established convergence guarantees for GANs. 
The work of \citep{pmlr-v119-lei20b,pmlr-v97-hsieh19b,mean-field-gan2020,Feizi2020,sun2020}
establish global convergence as described in Section~\ref{ss:contribution}.
\citet{cho2019} establish that the solution to the Wasserstein GAN is equivalent to PCA in the setup of learning a Gaussian distribution but do not make explicit guarantees on the training dynamics.
\citet{sanjabi2018} use a maximization oracle on a regularized Wassertstein distance to obtain an algorithm converging to stationary points, but did not provide any results relating to global optimality.

Although we do not make the connection formal, there is a large body of work establishing convergence for non-convex optimization problems with no spurious local minima solved with gradient descent \citep{lee2016,lee2019} and stochastic gradient descent \citep{ge15,jin17}.
The implication of having no spurious stationary points is that stochastic gradient descent finds a global minimum.

\subsection{Contribution}
\label{ss:contribution}
The key technical challenge of this work is the non-convexity of the loss function in the generator parameters, caused by the fact that the discriminator is nonlinear and non-convex in the input.
Prior work avoided this difficulty by using a linear discriminator \citep{pmlr-v119-lei20b} or by lifting the generator into the space of probability measures \citep{pmlr-v97-hsieh19b,sun2020,mean-field-gan2020}, also described as finding mixed Nash equilibria, but these are modifications not commonly used in the empirical training of GANs.
\citet{Feizi2020} seems to be the only exception, as they establish convergence guarantees for a WGAN with a linear generator and quadratic discriminator, but their setup is restricted to learning Gaussian distributions.
In contrast, we use a nonlinear discriminator and directly optimize the parameters without lifting to find mixed Nash equilibria (we find pure Nash equilibria), while using standard stochastic gradient ascent-descent.

To the best of our knowledge, our work is the first to use infinite-width analysis to establish theoretical guarantees for GANs with a nonlinear discriminator trained with stochastic gradient-type methods.
Our proof technique, distinct from the NTK or mean-field techniques, utilizes universal approximation theory and random feature learning to establish that there are no spurious stationary points. 
The only other prior work to use infinite-width analysis to study GANs was presented in \citep{mean-field-gan2020}, where the mean-field limit was used to establish guarantees on finding mixed Nash equilibria.

We point out that considering the NTK or mean-field limits of the generator and/or the discriminator networks does not resolve the non-convexity of the loss in the generator parameters.
We adopt the random feature learning setup, where the hidden layer features are fixed, and optimize only the output layers for both the generator and the discriminator.
Doing so allows us to focus on the key challenge of establishing guarantees on the optimization landscape despite the non-convexity.

\section{Problem setup}
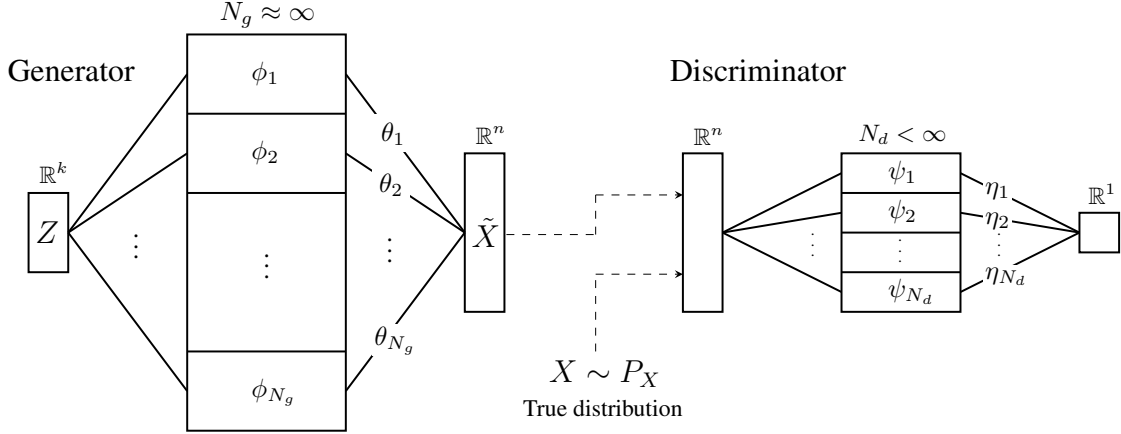
\begin{figure*}[h]
\begin{center}

\tikzset{every picture/.style={line width=0.75pt}} %set default line width to 0.75pt        

\begin{tikzpicture}[x=0.75pt,y=0.75pt,yscale=-1,xscale=1]
%uncomment if require: \path (0,372); %set diagram left start at 0, and has height of 372

%Shape: Rectangle [id:dp4496853127378415] 
\draw   (52,169) -- (72,169) -- (72,209) -- (52,209) -- cycle ;
%Shape: Rectangle [id:dp11101214951913962] 
\draw   (132,89) -- (212,89) -- (212,289) -- (132,289) -- cycle ;
%Straight Lines [id:da9027791658471072] 
\draw    (132,129) -- (212,129) ;
%Straight Lines [id:da9696315542359482] 
\draw    (132,169) -- (212,169) ;
%Straight Lines [id:da6770597335244719] 
\draw    (132,249) -- (212,249) ;
%Straight Lines [id:da9144440591168235] 
\draw    (72,189) -- (132,109) ;
%Straight Lines [id:da7877036514157425] 
\draw    (72,189) -- (132,149) ;
%Straight Lines [id:da7628216598962261] 
\draw    (72,189) -- (132,269) ;
%Shape: Rectangle [id:dp8673317231728346] 
\draw   (272,149) -- (292,149) -- (292,229) -- (272,229) -- cycle ;
%Straight Lines [id:da5064628854865401] 
\draw    (212,109) -- (272,189) ;
%Straight Lines [id:da5528915906990957] 
\draw    (212,149) -- (272,189) ;
%Straight Lines [id:da00008476346608210505] 
\draw    (212,269) -- (272,189) ;
%Shape: Rectangle [id:dp5748099231445511] 
\draw   (382,149) -- (402,149) -- (402,229) -- (382,229) -- cycle ;
%Shape: Rectangle [id:dp3020511709332834] 
\draw   (462,149) -- (522,149) -- (522,229) -- (462,229) -- cycle ;
% arrow from generated samples
\draw [->,>=stealth] [dashed] [line width=0.35pt]    (338,170) -- (382,170) ;
\draw  [dashed] [line width=0.35pt] (338,190) -- (338,170);
\draw  [dashed] [line width=0.35pt] (293,190) -- (338,190);

% arrow from true distribution
\draw [->,>=stealth] [dashed] [line width=0.35pt]    (338,210) -- (382,210) ;
\draw  [dashed] [line width=0.35pt] (338,249) -- (338,209.1);

%Straight Lines [id:da8409930751236117] 
\draw   (462,169) -- (522,169) ;
%Straight Lines [id:da21296981314393482] 
\draw    (462,189) -- (522,189) ;
%Straight Lines [id:da3369593282930441] 
\draw    (462,209) -- (522,209) ;
%Straight Lines [id:da9457805840526894] 
\draw    (402,189) -- (462,159) ;
%Straight Lines [id:da7475044853062827] 
\draw    (402,189) -- (462,179) ;
%Straight Lines [id:da05357268497038414] 
\draw    (402,189) -- (462,219) ;
%Straight Lines [id:da8676071674296757] 
\draw    (522,159) -- (582,189) ;
%Straight Lines [id:da1283811272306561] 
\draw    (522,179) -- (582,189) ;
%Straight Lines [id:da6616418769830756] 
\draw    (522,219) -- (582,189) ;
%Shape: Rectangle [id:dp4764435218034915] 
\draw   (582,179) -- (602,179) -- (602,199) -- (582,199) -- cycle ;
% %Shape: Ellipse [id:dp5796670042837602] 
% \draw   (332,279) .. controls (332,273.48) and (358.86,269) .. (392,269) .. controls (425.14,269) and (452,273.48) .. (452,279) .. controls (452,284.52) and (425.14,289) .. (392,289) .. controls (358.86,289) and (332,284.52) .. (332,279) -- cycle ;

% Generator Node
\draw (40,100) node [anchor=north west][inner sep=0.75pt]   [align=left] [font=\large] {Generator};
% Discriminator Node
\draw (374,100) node [anchor=north west][inner sep=0.75pt]   [align=left] [font=\large] {Discriminator};

% Text Node
\draw (56,152) node [anchor=north west][inner sep=0.75pt]  [font=\small] [align=left] {$\mathbb{R}^{k}$};
% Text Node
\draw (54,182) node [anchor=north west][inner sep=0.75pt] [font=\large]   [align=left] {$Z$};
% Text Node
\draw (147,71) node [anchor=north west][inner sep=0.75pt]   [align=left] {$N_{g} \approx \infty$};
% Text Node
\draw (163,101) node [anchor=north west][inner sep=0.75pt]   [align=left] {$\phi_{1}$};
% Text Node
\draw (163,141) node [anchor=north west][inner sep=0.75pt]   [align=left] {$\phi_{2}$};
% Text Node
\draw (163,261) node [anchor=north west][inner sep=0.75pt]   [align=left] {$\phi_{N_{g}}$};
% Text Node
\draw (169,190.4) node [anchor=north west][inner sep=0.75pt]    {$\vdots $};
% Text Node
\draw (103,180.4) node [anchor=north west][inner sep=0.75pt]    {$\vdots $};
% Text Node
\draw (274,180) node [anchor=north west][inner sep=0.75pt] [font=\large]  [align=left] {$\tilde{X}$};
% Text Node
\draw (226,129) node [anchor=north west][inner sep=0.75pt] [fill=white]  [align=left] [fill=white, minimum size=0.5cm] {$\theta_{1}$};
% Text Node
\draw (225,155) node [anchor=north west][inner sep=0.75pt] [fill=white]  [align=left] [fill=white, minimum size=0.5cm] {$\theta_{2}$};
% Text Node
\draw (225,235) node [anchor=north west][inner sep=0.75pt]  [fill=white] [align=left] [fill=white] {$\theta_{N_{g}}$};
% Text Node
\draw (276,134) node [anchor=north west][inner sep=0.75pt]  [font=\small] [align=left] {$\mathbb{R}^{n}$};
% Text Node
\draw (386,135) node [anchor=north west][inner sep=0.75pt]  [font=\small] [align=left] {$\mathbb{R}^{n}$};
% Text Node
% \draw (383,181) node [anchor=north west][inner sep=0.75pt] [font=\large]  [align=left] {$\tilde{X}$\\$X$};
% 5th vdots
\draw (489,183) node [anchor=north west][inner sep=0.75pt]  [font=\tiny]  {$\vdots $};
% Text Node
\draw (484,151) node [anchor=north west][inner sep=0.75pt]   [align=left] {$\psi_{1}$};
% Text Node
\draw (484,171) node [anchor=north west][inner sep=0.75pt]   [align=left] {$\psi_{2}$};
% Text Node
\draw (484,211) node [anchor=north west][inner sep=0.75pt]   [align=left] {$\psi_{N_{d}}$};
% 4th vdots
\draw (445,180) node [anchor=north west][inner sep=0.75pt]  [font=\tiny]  {$\vdots $};
% Text Node
\draw (585,163) node [anchor=north west][inner sep=0.75pt]  [font=\small] [align=left] {$\mathbb{R}^{1}$};
% Text Node
\draw (469,133) node [anchor=north west][inner sep=0.75pt]  [font=\small] [align=left] {$N_{d}<\infty$};
% eta_1
\draw (533,161) node [anchor=north west][inner sep=0.75pt] [fill=white, minimum size=0.35cm]  [align=left] {$\eta_{1}$};
% eta_2
\draw (533,176) node [anchor=north west][inner sep=0.75pt] [fill=white,minimum size=0.35cm]   [align=left] {$\eta_{2}$};
% eta_Nd
\draw (533,205) node [anchor=north west][inner sep=0.75pt] [fill=white]  [align=left] {$\eta_{N_{d}}$};
% 6th vdots
\draw (539,180) node [anchor=north west][inner sep=0.75pt]  [font=\tiny]  {$\vdots $};
% 3rd vdots
\draw (231,182.4) node [anchor=north west][inner sep=0.75pt]   {$\vdots $};
% Text Node
\draw (300,272) node [anchor=north west][inner sep=0.75pt]   [align=left] {{\small True distribution}};
\draw (313,252) node [anchor=north west][inner sep=0.75pt]   [align=left] [font=\large] {$X \sim P_{X}$};

\end{tikzpicture}

\end{center}
 \vspace{-1.2em}
\caption{Illustration of the generator and discriminator architectures.
$Z$ represents the latent variable, $\tilde{X}$ the generated samples, and $X$ the true samples.
Respectively, $\{\phi_i\}_{i=1}^{N_g}$ and $\{\psi_j\}_{j=1}^{N_d}$ are the generator and discriminator (post-activation) feature functions.
Respectively, $\{\theta_i\}_{i=1}^{N_g}$ and $\{\eta_j\}_{j=1}^{N_d}$ are the trainable parameters of the generator and discriminator networks.}
 \vspace{-0.5em}
\label{figure:architecture}    
\end{figure*}

%%% Definitions
%%%%%%%%%%%%%%%
We consider a WGAN whose generator and the discriminator are two-layer networks as illustrated in Figure~\ref{figure:architecture}.

Let $X\in \fR^n$ be a random vector with a true (target) distribution $P_X$.
%$\in \cP(\fR^n)$, where $\cP(\fR^n)$ is the set of probability measures on $\fR^n$.
Let $Z\in \fR^k$ be a continuous random vector from the latent space satisfying the following assumption.
% with distribution $P_Z\in \cP(\fR^k)$
% and $\mathrm{supp}(P_Z)=\fR^k$, such as a unit Gaussian. \noa{need more condition to bound the pdf}

\setlength{\leftskip}{0.2in}
\setlength{\rightskip}{0in}

\hypertarget{cond:al}{\textbf{(AL)}}
The latent vector $Z\in \fR^k$ has a Lipschitz continuous probability density function $q_Z(z)$ satisfying $q_Z(z)>0$ for all $z\in\fR^k$.

\setlength{\leftskip}{0in}
\setlength{\rightskip}{0in}

The standard Gaussian is a possible choice satisfying \hyperlink{cond:al}{(AL)}.

%%% Generator Class
\subsection{Generator Class}
Let $\cG = \{\phi(\cdot; \kappa) \,|\, \kappa\in \fR^p\}$, where $\phi(\cdot;\kappa)\colon\fR^k\rightarrow \fR^n$, be a collection of generator feature functions satisfying the following assumption.
%parameterized by $\kappa\in \fR^p$ is a continuous functions.

\setlength{\leftskip}{0.2in}
\setlength{\rightskip}{0in}

\hypertarget{cond:AG}{\textbf{(AG)}}
All generator feature functions $\phi\in \cG$ are of form $\phi(z;\kappa)=\sigma_\mathrm{g}(\kappa_wz+\kappa_b)$,
where $\kappa=(\kappa_w,\kappa_b)\in\fR^{n\times k}\times \fR^n$, and $\sigma_\mathrm{g}\colon\fR\rightarrow\fR$ is a bounded continuous activation function
satisfying $\lim_{r\rightarrow-\infty}\sigma_\mathrm{g}(r)<\lim_{r\rightarrow \infty}\sigma_\mathrm{g}(r)$.
(So $p=nk+n$.)

\setlength{\leftskip}{0in}
\setlength{\rightskip}{0in}

\begin{definition}[Generator class, finite width]\label{def:DiscreteGenerators}
Consider the generator feature functions $\phi_1, \ldots, \phi_{N_g}\in\cG$, where $1\le N_g<\infty$.
For $\theta\in \fR^{N_g}$, let
\[
g_{\theta}(z) = \sum_{i=1}^{N_g} \theta_i\phi_i(z).
\]
Write 
\[
\mathrm{span} (\{\phi_i\}_{i=1}^{N_g})= \left\{g_\theta \,|\, \theta \in \fR^{N_g}\right\}
\]
for the class of generators constructed from the feature functions in $\{\phi_i\}_{i=1}^{N_g}$.
% % \theta=& \left[\theta_1, \ldots, \theta_{N_g}\right]^\intercal\\
% \phi_i(z) =& \phi(z;\kappa_i) \quad\quad\mbox{ $1\leq i\leq N_g$}.
% \end{align*}
\end{definition}

Note that there exists $\kappa_i\in \fR^p$ such that $\phi_i(z) = \phi(z;\kappa_i)$ for $1\le i\le N_g$.
% XXX The generator class $\mathrm{span}(\{\phi_i\}_{i=1}^{N_g})$ is the span of the feature functions $\{\phi_i\}_{i=1}^{N_g}$. XXX
We can view the generator $g_{\theta}$ as a two-layer network, where $\phi(z;\kappa)=\sigma_\mathrm{g}(\kappa_wz+\kappa_b)$ represents the post-activation values of the hidden layer.

\begin{definition}[Generator class, infinite width] \label{def:ContinuousGenerators}
For $ \theta \in \cM(\fR^p)$, where $\cM(\fR^p)$ is the set of measures on $\fR^p$ with finite total mass, let
\[
g_{\theta}(z) = \int \phi(z;\kappa) \, d\theta(\kappa).
\]
Write
\[
\overline{\mathrm{span}} (\cG)
= \left\{g_{\theta}(z)\,|\, \theta \in \cM(\fR^p)\right\}
\]
for the class of infinite-width generators constructed from the feature functions in $\cG$.
\end{definition}

We assume the class of generator feature functions $\cG$ satisfies the following universality property.

\setlength{\leftskip}{0.2in}
\setlength{\rightskip}{0in}

\hypertarget{cond:univ_apx}{\textbf{(Universal approximation property)}}
For any function $f\colon\fR^k \rightarrow \fR^n$ such that $\EZ{\|f(Z)\|_2}<\infty$ and $\varepsilon>0$, there exists $\theta_{\varepsilon}\in\cM(\fR^p)$ such that
\[
\EZ{ \left\|g_{\theta_\varepsilon}(Z) - f(Z) \right\|_2 } < \varepsilon.
\]
\setlength{\leftskip}{0in}
\setlength{\rightskip}{0in}

This assumption holds quite generally.
In particular, the following lemma holds as a consequence of \citep{HORNIK1991251}.
% as \citep[Theorem~1]{HORNIK1991251} establishes.
% To clarify, \citet{HORNIK1991251} presents a more general result that implies Lemma~\ref{lem:universal}.

\begin{lemma}
\label{lem:universal}
\hyperlink{cond:AG}{(AG)} implies $\cG$ satisfies the \hyperlink{cond:univ_apx}{(Universal approximation property)}.
%holds with $\phi(z;\kappa)=\sigma_\mathrm{g}(\kappa_wz+\kappa_b)$.
% Let $\sigma\colon\fR\rightarrow\fR$ be bounded and nonconstant and $P$ be a finite measure.
% Then for any integrable $f\colon\fR^n\rightarrow\fR^n$ and $\varepsilon>0$,
% there exists large enough $N$ and $\{(w_i,A_i,b_i)\}_{k=1}^N$ such that 
% \begin{align}
% \int_{\fR^k}\left\|\sum_{i=1}^Nw_i\sigma (A_iz-b_i)-f(z)\right\|_2\,dP(z)
% <\varepsilon.
% \end{align}
\end{lemma}

% We detail the translation of Hornik
% We perform the precise translation in the appendix, in Section~\ref{appendix:univ-apx}.

In functional analytical terms, \hyperlink{cond:univ_apx}{(Universal approximation property)} states that $\overline{\mathrm{span}} (\cG)$ is dense in $L^1(q_Z(z)\,dz;\fR^n)$.
Later in the proof of Theorem~\ref{thm:infinite}, we instead use the following dual characterization of denseness.
% Lemma~\ref{lem:dual-universal} is the dual characterization of denseness.

\begin{lemma}
\label{lem:dual-universal}
Assume \hyperlink{cond:al}{(AL)} and \hyperlink{cond:univ_apx}{(Universal approximation property)}.
If a bounded continuous function $h\colon \fR^k\rightarrow \fR^n$ satisfies
% If a continuous function $h\colon \fR^k\rightarrow \fR$ satisfies
% If $\mu$ is a Radon measure such that $d\mu \ll q_Zdz$ and 
\[
\EZ{ \phi^\intercal(Z)h(Z)}=0\qquad \forall \phi\in \cG,
\]
then $h(z)=0$ for all $z\in\fR^k$.
\end{lemma}
% We provide the proof in the appendix.

%%% Discriminator Class
\subsection{Discriminator Class}
Let $\cD=\{\psi_1, \ldots, \psi_{N_d}\}$ be a class of discriminator feature functions $\psi_j\colon\fR^n \rightarrow \fR$ for $1\leq j\leq N_d$ satisfying the following assumption.

\setlength{\leftskip}{0.2in}
\setlength{\rightskip}{0in}

\hypertarget{cond:AD}{\textbf{(AD)}} For all $1\leq j\leq N_d$, the discriminator feature functions are of form $\psi_j(x) = \sigma(a^\intercal_jx+b_j)$ for some $a_j\in\fR^n$ and $b_j\in\fR$.
The twice differentiable activation function $\sigma$ satisfies $\sigma'(x)>0$ for all $x\in \fR$ and $\sup_{x\in\fR}|\sigma(x)|+|\sigma'(x)|+|\sigma''(x)|<\infty$.
The weights $a_1, \ldots, a_{N_d}$ and biases $b_1, \ldots, b_{N_d}$ are sampled (IID) from a distribution with a probability density function.

\setlength{\leftskip}{0in}
\setlength{\rightskip}{0in}

The sigmoid or $\tanh$ activation functions for $\sigma$ and the standard Gaussian for the distribution of $a_1, \ldots, a_{N_d}$ and $b_1, \ldots, b_{N_d}$ are possible choices satisfying \hyperlink{cond:AD}{(AD)}.
To clarify with measure-theoretic terms, we are assuming that $a_1, \ldots, a_{N_d}$ and $b_1, \ldots, b_{N_d}$ are sampled from a probability distribution that is absolutely continuous with respect to the Lebesgue measure.

\begin{definition}[Discriminator Class]
For $\eta\in \fR^{N_d}$, let
\[
\Psi(x) = (\psi_1(x), \ldots, \psi_{N_d}(x))\in \fR^{N_d}
\]
and
\[
f_{\eta}(x) = \sum_{j=1}^{N_d}\eta_j\psi_j(x)=\eta^\intercal\Psi(x).
\]
Write
\begin{align*}
\mathrm{span}(\cD)
=& \left\{f_{\eta}\,|\, \eta\in \fR^{N_d}\right\}
\end{align*}
for the class of discriminators constructed from the feature functions in $\cD$.
% \begin{align*}
% \eta =& \left[\eta_1, \ldots, \eta_{N_d}\right]^\intercal\\
% \Psi(x) =& \left[\psi_1(x), \ldots, \psi_{N_d}(x)\right]^\intercal
% \end{align*}
\end{definition}

In contrast with the generators, we only consider finite-width discriminators.

%%% Two-Layer Generative Adversarial Networks 
%%%%%%%%%%%%%%%
\subsection{Adversarial training with stochastic gradients}\label{sec:twolayergan}
% The goal of the generator is to find $\theta\in\mu(\fR^p)$ where $g_{\theta}(Z)$ has a similar distribution with the true distribution $P_X$.
% The goal of the discriminator is to find $\eta\in \fR^{N_d}$ where $f_{\eta}$ distinguishes between the generated output $g_\theta(Z)$ and the sample from the true distribution $X$.
% \noa{The discriminator assigns the higher value if it thinks the sample is from true distribution.}
Consider the loss
\begin{align*}
L(\theta, \eta) &= \EX{f_{\eta}(X)} - \EZ{f_{\eta}(g_{\theta}(Z))} -\frac{1}{2}\|\eta\|_2^2\\
&= \EX{\eta^\intercal \Psi(X)} - \EZ{\eta^\intercal \Psi(g_{\theta}(Z))} - \frac{1}{2}\|\eta\|_2^2.
\end{align*}
This is a variant of the WGAN loss with the Lipschitz constraint on the discriminator replaced with an explicit regularizer.
This loss and regularizer were also considered in \citep{pmlr-v119-lei20b}.

We train the two networks adversarially by solving the minimax problem
\[
\underset{\theta}{\mbox{minimize}}\,\,
\underset{\eta}{\mbox{maximize}}\,\, 
L(\theta, \eta)
\]
using stochastic gradient ascent-descent\footnote{In the infinite-width case, where $\theta\in \cM(\fR^p)$, the stochastic gradient method we describe is not well defined as $\nabla_\theta$, if formally defined, is not an element of $ \cM(\fR^p)$.
For a rigorous treatment of analogs of gradient descent in $\cM(\fR^p)$, see \citep{mei2019mean,chizat2019sparse} and reference therein.
In this work, we apply stochastic gradient ascent-descent only in the finite-width setup, but we analyze stationary points for both the finite and infinite setups.
}
\begin{align*}
\gamma^{t}_\eta&=\Psi(X^t)- \Psi(g_{\theta}(Z^t_1))-\eta^t\\
\eta^{t+1} &= \eta^t +\gamma^{t}_\eta\\
\gamma^{t}_\theta&= (D_\theta\Psi(g_{\theta}(Z_2^t)))^\intercal \eta^{t+1}\\
\theta^{t+1} &=  \theta^t - \alpha \gamma^{t}_\theta
\end{align*}
for $t=0,1,\dots$, 
where $X^t\sim P_X$, $Z_1^t\sim P_Z$, and $Z_2^t\sim P_Z$ are independent. We fix the maximization stepsize to $1$ while letting the minimization stepsize be $\alpha>0$.
Note that $\gamma^{t}_\eta$ and $\gamma^{t}_\theta$ are stochastic gradients in the sense that $\E {\gamma^{t}_\eta}=\nabla_{\eta} L(\theta^t, \eta^t)$
and $\E {\gamma^{t}_\theta}=\nabla_{\theta} L(\theta^t, \eta^{t+1})$.
We can also form $\gamma^{t}_\eta$ and $\gamma^{t}_\theta$ with batches.
To clarify,
\begin{align*}
&D_\theta\Psi(g_{\theta}(Z))
=
\begin{bmatrix}
(\nabla_\theta \left(\psi_1(g_{\theta}(Z))\right))^\intercal\\
\vdots\\
(\nabla_\theta \left(\psi_{N_d}(g_{\theta}(Z))\right))^\intercal
\end{bmatrix}\\
&=
\begin{bmatrix}
(\nabla_x \psi_1(g_{\theta}(Z)))^\intercal
\\
\vdots
\\
(\nabla_x \psi_{N_d}(g_{\theta}(Z)))^\intercal
\end{bmatrix}
\begin{bmatrix}
\phi_1(Z)&\cdots&\phi_{N_g}(Z)
\end{bmatrix}.
\end{align*}

% In the finite width case, where $\theta\in \fR^{N_g}$, this update is well-defined.

The minimax problem is equivalent to the minimization problem 
\[
\inf_{\theta}\sup_{\eta}L(\theta, \eta) = \inf_{\theta} J(\theta), 
\]
where 
\begin{align*}
J(\theta) &\stackrel{\Delta}{=}
\sup_{\eta} L(\theta, \eta)\\
&= \frac{1}{2}\left\|\EX{\Psi(X)} -\EZ{\Psi(g_{\theta}(Z))}\right\|_2^2.
\end{align*}
Interestingly, stochastic gradient ascent-descent applied to $L(\theta,\eta)$ is equivalent to stochastic gradient descent applied to $J(\theta)$:
eliminate the $\eta$-variable in the iteration to get
\begin{align*}
\theta^{t+1} &=  \theta^t - \alpha(D_\theta\Psi(g_{\theta}(Z_2^t)))^\intercal (\Psi(X^t)- \Psi(g_{\theta}(Z^t_1)))
\end{align*}
and note
\begin{align*}
&\E{D_\theta\Psi(g_{\theta}(Z_2^t))^\intercal (\Psi(X^t)- \Psi(g_{\theta}(Z^t_1)))}\\
&\qquad=\EZ{(D_\theta\Psi(g_{\theta}(Z))^\intercal}(\EX{\Psi(X)}- \EZ{\Psi(g_{\theta}(Z))})\\
&\qquad=\nabla_\theta J(\theta).
\end{align*}
In the following sections, we show that $J(\theta)$ has no spurious stationary points under suitable conditions.

Finally, we introduce the notation
\[
r(\theta) = \EX{\Psi(X)} - \EZ{\Psi(g_{\theta}(Z))},
\]
i.e., $r_j(\theta) = \EX{\psi_j(X)} - \EZ{\psi_j(g_{\theta}(Z))}$ for $1\le j\le N_d$.
This allows us to write $J(\theta) = \frac{1}{2}\|r(\theta)\|_2^2$.

\begin{figure*}[h]
\begin{center}

\begin{tikzpicture}[x=0.75pt,y=0.75pt,yscale=-1,xscale=1]
%uncomment if require: \path (0,435); %set diagram left start at 0, and has height of 435

%Rounded Rect [id:dp5680888442234742] 
\draw   (70,83.64) .. controls (70,80.53) and (72.53,78) .. (75.64,78) -- (116.49,78) .. controls (119.61,78) and (122.13,80.53) .. (122.13,83.64) -- (122.13,100.56) .. controls (122.13,103.67) and (119.61,106.2) .. (116.49,106.2) -- (75.64,106.2) .. controls (72.53,106.2) and (70,103.67) .. (70,100.56) -- cycle ;
%Straight Lines [id:da6051282329239198] 
\draw [->,>=stealth] (122,91) -- (191,91) ;
%Rounded Rect [id:dp5716432251502555] 
\draw   (191,82.64) .. controls (191,79.53) and (193.53,77) .. (196.64,77) -- (237.49,77) .. controls (240.61,77) and (243.13,79.53) .. (243.13,82.64) -- (243.13,99.56) .. controls (243.13,102.67) and (240.61,105.2) .. (237.49,105.2) -- (196.64,105.2) .. controls (193.53,105.2) and (191,102.67) .. (191,99.56) -- cycle ;
%Rounded Rect [id:dp6998852172786467] 
\draw   (312,75.89) .. controls (312,71.01) and (315.95,67.06) .. (320.83,67.06) -- (394.3,67.06) .. controls (399.18,67.06) and (403.13,71.01) .. (403.13,75.89) -- (403.13,102.37) .. controls (403.13,107.25) and (399.18,111.2) .. (394.3,111.2) -- (320.83,111.2) .. controls (315.95,111.2) and (312,107.25) .. (312,102.37) -- cycle ;
%Straight Lines [id:da002045289603319622] 
\draw  [->,>=stealth]  (243,86) -- (312,85.73) ;

%Straight Lines [id:da7028025668982063] 
\draw    (122,146) -- (250.24,145.72) ;
%Straight Lines [id:da03216316928018714] 
\draw    (250.24,91.72) -- (250.24,145.72) ;

%Straight Lines [id:da1398940582373478] 
\draw  [->,>=stealth]  (250.24,91.72) -- (312,91.72) ;

% No spurious local min N_g = infty, Nd<=n 
\draw   (477,68.18) .. controls (477,61.14) and (482.71,55.43) .. (489.75,55.43) -- (580.5,55.43) .. controls (587.55,55.43) and (593.26,61.14) .. (593.26,68.18) -- (593.26,106.45) .. controls (593.26,113.49) and (586.55,119.2) .. (579.5,119.2) -- (489.75,119.2) .. controls (482.71,119.2) and (477,113.49) .. (477,106.45) -- cycle ;
%Straight Lines [id:da44424702263366056] 
\draw   [->,>=stealth] (403,83) -- (477,82.44) ;

%Straight Lines [id:da5719204626477192] 
\draw  [->,>=stealth]  (412.26,88.43) -- (477,88.43) ;

%Rounded Rect [id:dp035364856417057444] 
\draw   (350,186.64) .. controls (350,183.53) and (352.53,181) .. (355.64,181) -- (396.49,181) .. controls (399.61,181) and (402.13,183.53) .. (402.13,186.64) -- (402.13,203.56) .. controls (402.13,206.67) and (399.61,209.2) .. (396.49,209.2) -- (355.64,209.2) .. controls (352.53,209.2) and (350,206.67) .. (350,203.56) -- cycle ;
%Rounded Rect [id:dp28486523941759057] 
\draw   (229,186.64) .. controls (229,183.53) and (231.53,181) .. (234.64,181) -- (275.49,181) .. controls (278.61,181) and (281.13,183.53) .. (281.13,186.64) -- (281.13,203.56) .. controls (281.13,206.67) and (278.61,209.2) .. (275.49,209.2) -- (234.64,209.2) .. controls (231.53,209.2) and (229,206.67) .. (229,203.56) -- cycle ;
%Straight Lines [id:da10545174550409886] 
\draw    (412.26,88.43) -- (412.26,194.06) ;
%Straight Lines [id:da7713748921723123] 
\draw    (412.26,194.06) -- (402.26,194.06) ;
%Rounded Rect [id:dp7388164123058569] 
\draw   (70,137.64) .. controls (70,134.53) and (72.53,132) .. (75.64,132) -- (116.49,132) .. controls (119.61,132) and (122.13,134.53) .. (122.13,137.64) -- (122.13,154.56) .. controls (122.13,157.67) and (119.61,160.2) .. (116.49,160.2) -- (75.64,160.2) .. controls (72.53,160.2) and (70,157.67) .. (70,154.56) -- cycle ;
%Rounded Rect [id:dp5059291031025284] 
\draw   (351,241.64) .. controls (351,238.53) and (353.53,236) .. (356.64,236) -- (397.49,236) .. controls (400.61,236) and (403.13,238.53) .. (403.13,241.64) -- (403.13,258.56) .. controls (403.13,261.67) and (400.61,264.2) .. (397.49,264.2) -- (356.64,264.2) .. controls (353.53,264.2) and (351,261.67) .. (351,258.56) -- cycle ;
%Rounded Rect [id:dp024946346343098114] 
\draw   (204,238) .. controls (204,233.58) and (203.58,230) .. (212,230) -- (266,230) .. controls (278.42,230) and (282,233.58) .. (282,238) -- (282,262) .. controls (282,266.42) and (278.42,270) .. (274,270) -- (220,270) .. controls (207.58,270) and (204,266.42) .. (204,262) -- cycle ;
%Straight Lines [id:da649494918226394] 
\draw   [->,>=stealth] (281,190) -- (350,189.73) ;

%Straight Lines [id:da4103977844466207] 
\draw    (281.26,197.06) -- (313,197.06) ;
%Straight Lines [id:da585711673598764] 
\draw    (313,197.06) -- (313,246.06) ;
%Straight Lines [id:da6397813292598367] 
\draw   [->,>=stealth] (282,252) -- (350.5,252) ;

%Straight Lines [id:da6386691633146309] 
\draw   [->,>=stealth] (313,246) -- (350.5,246) ;

% No spurious local min Ng = infty, Nd <= n
\draw   (478,231.18) .. controls (478,224.14) and (483.71,218.43) .. (490.75,218.43) -- (580.5,218.43) .. controls (587.55,218.43) and (593.26,224.14) .. (593.26,231.18) -- (593.26,269.45) .. controls (593.26,276.49) and (586.55,282.2) .. (579.5,282.2) -- (490.75,282.2) .. controls (483.71,282.2) and (478,276.49) .. (478,269.45) -- cycle ;
%Straight Lines [id:da8595775136496207] 
\draw    (403,96) -- (440,96) ;
%Straight Lines [id:da14443444854166798] 
\draw    (440,96) -- (440,247.06) ;
%Straight Lines [id:da7522698077802401] 
\draw   [->,>=stealth] (440,247.06) -- (478,247.06) ;
%Straight Lines [id:da40573843417186883] 
\draw  [->,>=stealth]  (403,253) -- (478,253) ;

% Text Node
\draw (96.07,92.1) node   [align=left] {(AG)};
% Text Node
\draw (217.07,91.1) node   [align=left] {(UAP)};
% Text Node
\draw (357.57,89.13) node   [align=center] {denseness of\\ $\cG$};
% Text Node
\draw (535.63,87.31) node   [align=center] {no spurious\\local min\\$N_{g}=\infty$, $N_{d}\leq n$};
% Text Node
\draw (255.07,195.1) node   [align=left] {(AD)};
% Text Node
\draw (376.07,195.1) node   [align=left] {(JKP)};
% Text Node
\draw (96.07,146.1) node   [align=left] {(AL)};
% Text Node
\draw (377.07,250.1) node   [align=left] {(JKB)};
% Text Node
\draw (242,250) node   [align=left] {$\sigma$ = sigmoid\\ \ \ \ \ \ \ \ tanh};
% Text Node
\draw (535.63,250.31) node  [align=center] {no spurious\\local min\\$N_{g}=\infty$, $N_{d}>n$};
% Text Node
\draw (138,76) node [anchor=north west][inner sep=0.75pt]   [align=left] {Lem~\ref{lem:universal}};
% Text Node
\draw (260,70) node [anchor=north west][inner sep=0.75pt]   [align=left] {Lem~\ref{lem:dual-universal}};
% Text Node
\draw (423,68) node [anchor=north west][inner sep=0.75pt]   [align=left] {Thm~\ref{thm:infinite}};
% Text Node
\draw (295,175) node [anchor=north west][inner sep=0.75pt]   [align=left] {Lem~\ref{lem:infinite small justification}};
% Text Node
\draw (295,255) node [anchor=north west][inner sep=0.75pt]   [align=left] {Lem~\ref{lem:infinite large justification}};
% Text Node
\draw (423,256) node [anchor=north west][inner sep=0.75pt]   [align=left] {Thm~\ref{thm:infinite_large}};

\end{tikzpicture}

\end{center}
 \vspace{-1.2em}
\caption{The diagrammatic summary of proofs in Section \ref{sec:infinite}.
We denote \hyperlink{cond:univ_apx}{(Universal approximation property)} by (UAP),
\hyperlink{cond:kernel_point}{(Jacobian kernel point condition)} by (JKP),
and \hyperlink{cond:kernel_ball}{(Jacobian kernel ball condition)} by (JKB).
}
 \vspace{-0.5em}
\label{figure:asumptions1}    
\end{figure*}

%%%%%%%%%%%%%%%%%%%%%%%%%%%%%%%%%%%%%%%%%%%%%%%%%%
%%% Infinite Width Generator
%%%%%%%%%%%%%%%%%%%%%%%%%%%%%%%%%%%%%%%%%%%%%%%%%%
\section{Infinite-width generator}\label{sec:infinite}
Consider a GAN with a two-layer \emph{infinite-width} generator $g_\theta\in \overline{\mathrm{span}} (\cG)$ and a two-layer finite-width discriminator $f_{\eta}(x)\in \mathrm{span}(\cD)$.
In this section, we show that under suitable conditions, $J(\theta)$ has no spurious stationary points, i.e., a stationary point of $J(\theta)$ is necessarily a global minimum.

We say $\theta_\mathrm{s}$ is a stationary point of $J$ if $J(\theta_\mathrm{s}+\lambda\mu)$, as a function of $\lambda\in\fR$, is differentiable and has zero gradient at $\lambda=0$ for any $\mu\in\cM(\fR^p)$.
% Under this definition, a local minimum is a stationary point.

%%%% Small discriminator
%%%%%%%%%%%%%%%%%%%%%%%
\subsection{Small discriminator $(N_d\leq n)$}\label{subsec:small discriminator}
We first consider the case where the discriminator has width $N_d\leq n$. 
Consider the following condition.

% XXX
% Consider the class of infinitely wide generator functions $\overline{\mathrm{span}}(\cG)$ as defined in Definition \ref{def:ContinuousGenerators}.
% . XXX

\setlength{\leftskip}{0.2in}
\setlength{\rightskip}{0in}

\hypertarget{cond:kernel_point}{\textbf{(Jacobian kernel point condition)}}
The Jacobian
\[
D\Psi(x)^\intercal = 
\begin{bmatrix}
\nabla\psi_1(x) &
\nabla\psi_2(x) & 
\cdots &  \nabla\psi_{N_d}(x)
\end{bmatrix}\]
satisfies $\ker (D\Psi(x)^\intercal) = \{\bzero{N_d}\}$ for all $x\in \fR^n$.

\setlength{\leftskip}{0in}
\setlength{\rightskip}{0in}

We can interpret \hyperlink{cond:kernel_point}{(Jacobian kernel point condition)} to imply that there is no redundancy in the discriminator feature functions $\psi_1,\dots,\psi_{N_d}$.
This condition holds quite generally when sigmoidal activation functions are used, as characterized by the following lemma.

\begin{lemma}\label{lem:infinite small justification}
\hyperlink{cond:AD}{(AD)} implies \hyperlink{cond:kernel_point}{(Jacobian kernel point condition)} with probability $1$.
\end{lemma}

\begin{proof}
Since $\nabla\psi_j(x) = a_j \sigma'(a_j^\intercal x+b_j)$, we have
\begin{align*}
&D\Psi(x)^\intercal
=\begin{bmatrix}
\nabla\psi_1(x)&
\cdots&
\nabla\psi_{N_d}(x)
\end{bmatrix}\\
=& \begin{bmatrix}
a_1& \cdots & a_{N_d}
\end{bmatrix}
\mbox{diag}(\sigma'(a_1^\intercal x+b_1), \ldots, \sigma'(a_{N_d}^\intercal x + b_{N_d})).
\end{align*}
By \hyperlink{cond:AD}{(AD)},
$\begin{bmatrix} a_1& \cdots & a_{N_d} \end{bmatrix}$ has full rank with probability 1,
and therefore $D\Psi(x)^\intercal$ has full rank.
\end{proof}

We are now ready to state and prove the first main result of this work:
our GAN with an infinite-width generator has no spurious stationary points.

\begin{theorem}\label{thm:infinite}
Assume \hyperlink{cond:al}{(AL)}, \hyperlink{cond:AG}{(AG)}, and \hyperlink{cond:AD}{(AD)}.
Then the following statement holds%
\footnote{The randomness comes from the random generation of $\psi_j$'s described in \hyperlink{cond:AD}{(AD)} and is unrelated to randomness of SGD.
Once $\psi_j$'s have been generated and the \hyperlink{cond:kernel_point}{(Jacobian kernel point condition)} holds by Lemma~\ref{lem:infinite small justification}, the conclusion of Theorem~\ref{thm:infinite} holds without further probabilistic quantifiers.} 
with probability $1$: any stationary point $\theta_\mathrm{s}$ satisfies $J(\theta_\mathrm{s})=0$.
% , and thus there are no local minima.
\end{theorem}

\begin{proof}
Let $\theta_\mathrm{s}$ be a stationary point of $J$.
Then, for any $\mu \in \cM(\fR^p)$,
\[
\left.\frac{\partial}{\partial \lambda} J(\theta_\mathrm{s}+\lambda \mu)\right|_{\lambda=0} = 0.
\]
Since $g_{\theta_\mathrm{s}+\lambda\mu} = g_{\theta_\mathrm{s}} + \lambda g_{\mu}$, 
\begin{align*}
&\frac{\partial}{\partial \lambda} J(\theta_\mathrm{s}+\lambda \mu)
=\frac{\partial}{\partial\lambda}\frac{1}{2} \|r(\theta_\mathrm{s}+\lambda\mu)\|_2^2\\
&= - r(\theta_\mathrm{s}+\lambda\mu)^\intercal  \EZ{\frac{\partial}{\partial\lambda}\Psi(g_{\theta_\mathrm{s}+\lambda\mu}(Z))}\\
&= - r(\theta_\mathrm{s}+\lambda\mu)^\intercal \EZ{D\Psi(g_{\theta_\mathrm{s}+\lambda\mu}(Z))^\intercal g_{\mu}(Z)}\\
&= - \EZ{\sum_{j=1}^{N_d} r_j(\theta_\mathrm{s}+\lambda\mu)
\nabla\psi_j(g_{\theta_\mathrm{s}+\lambda\mu}(Z))^\intercal g_{\mu}(Z)}.
\end{align*}
Thus, for all $\mu\in\cM(\fR^p)$,
\[
\EZ{\sum_{j=1}^{N_d} r_j(\theta_\mathrm{s})
\nabla\psi_j(g_{\theta_\mathrm{s}}(Z))^\intercal g_{\mu}(Z)}
=0.
\]
By Lemmas~\ref{lem:universal} and \ref{lem:dual-universal}, 
\begin{equation}
\sum_{j=1}^{N_d} r_j(\theta_\mathrm{s})  \nabla\psi_j(g_{\theta_\mathrm{s}}(z)) = 0
\label{thm:infinite-key}
\end{equation}
for all $z\in\fR^k$.
Thus,
\begin{align*}
\begin{bmatrix}
\nabla\psi_1(g_{\theta_\mathrm{s}}(z)) &
\cdots&
\nabla\psi_{N_d}(g_{\theta_\mathrm{s}}(z))
\end{bmatrix}
% \times
\begin{bmatrix}
r_1(\theta_\mathrm{s})\\
\vdots\\
r_{N_d}(\theta_\mathrm{s})
\end{bmatrix}
= \bzero{n}.
\end{align*}
By Lemma~\ref{lem:infinite small justification}, the \hyperlink{cond:kernel_point}{(Jacobian kernel point condition)} holds with probability 1. Therefore, $r(\theta_\mathrm{s}) = 0$ and we conclude $J(\theta_\mathrm{s}) = 0$.
\end{proof}

\begin{figure*}[h]
\begin{center}
\begin{tikzpicture}[x=0.75pt,y=0.75pt,yscale=-1,xscale=1]
%uncomment if require: \path (0,435); %set diagram left start at 0, and has height of 435

%Rounded Rect [id:dp5680888442234742] 
\draw   (69,83.64) .. controls (69,80.53) and (71.53,78) .. (74.64,78) -- (115.49,78) .. controls (118.61,78) and (121.13,80.53) .. (121.13,83.64) -- (121.13,100.56) .. controls (121.13,103.67) and (118.61,106.2) .. (115.49,106.2) -- (74.64,106.2) .. controls (71.53,106.2) and (69,103.67) .. (69,100.56) -- cycle ;
%Rounded Rect [id:dp7388164123058569] 
\draw   (69,137.64) .. controls (69,134.53) and (71.53,132) .. (74.64,132) -- (115.49,132) .. controls (118.61,132) and (121.13,134.53) .. (121.13,137.64) -- (121.13,154.56) .. controls (121.13,157.67) and (118.61,160.2) .. (115.49,160.2) -- (74.64,160.2) .. controls (71.53,160.2) and (69,157.67) .. (69,154.56) -- cycle ;
%Rounded Rect [id:dp435460861099068] 
\draw   (69,201.64) .. controls (69,198.53) and (71.53,196) .. (74.64,196) -- (115.49,196) .. controls (118.61,196) and (121.13,198.53) .. (121.13,201.64) -- (121.13,218.56) .. controls (121.13,221.67) and (118.61,224.2) .. (115.49,224.2) -- (74.64,224.2) .. controls (71.53,224.2) and (69,221.67) .. (69,218.56) -- cycle ;
%Rounded Rect [id:dp820970573405919] 
\draw   (185,104.77) .. controls (185,99.47) and (189.3,95.17) .. (194.6,95.17) -- (308.66,95.17) .. controls (313.96,95.17) and (318.26,99.47) .. (318.26,104.77) -- (318.26,133.57) .. controls (318.26,138.87) and (313.96,143.17) .. (308.66,143.17) -- (194.6,143.17) .. controls (189.3,143.17) and (185,138.87) .. (185,133.57) -- cycle ;
%Straight Lines [id:da06384025780433222] 
\draw    (121.26,91.17) -- (130.26,91.17) ;
%Straight Lines [id:da6850550181566866] 
\draw    (121,146.5) -- (130.26,146.38) ;
%Rounded Rect [id:dp05288657797555518] 
\draw   (400,100.17) .. controls (400,89.68) and (408.51,81.17) .. (419,81.17) -- (526.26,81.17) .. controls (536.75,81.17) and (545.26,89.68) .. (545.26,100.17) -- (545.26,142.17) .. controls (545.26,152.66) and (536.75,161.17) .. (526.26,161.17) -- (419,161.17) .. controls (408.51,161.17) and (400,152.66) .. (400,142.17) -- cycle ;
%Straight Lines [id:da19865358027688695] 
\draw    (130.26,91.17) -- (130.26,116.17) ;
%Straight Lines [id:da12615834173060692] 
\draw   [->,>=stealth] (130.26,116.17) -- (185.26,116.17) ;
%Straight Lines [id:da9717283674531869] 
\draw    (130.26,122.17) -- (130.26,146.17) ;
%Straight Lines [id:da19497949914195045] 
\draw   [->,>=stealth] (130.26,122.17) -- (185.26,122.17) ;
%Straight Lines [id:da2785531131547738] 
\draw   [->,>=stealth] (318.26,116.17) -- (400.26,116.17) ;
%Rounded Rect [id:dp9907739300856802] 
\draw   (266,200.64) .. controls (266,197.53) and (268.53,195) .. (271.64,195) -- (312.49,195) .. controls (315.61,195) and (318.13,197.53) .. (318.13,200.64) -- (318.13,217.56) .. controls (318.13,220.67) and (315.61,223.2) .. (312.49,223.2) -- (271.64,223.2) .. controls (268.53,223.2) and (266,220.67) .. (266,217.56) -- cycle ;
%Straight Lines [id:da011735193043483694] 
\draw  [->,>=stealth]  (121,209) -- (266.26,209) ;
%Straight Lines [id:da9175576883835357] 
\draw    (318.26,209.17) -- (326.26,209.17) ;
%Straight Lines [id:da90628644446669] 
\draw    (326.26,122.17) -- (326.26,209.17) ;
%Straight Lines [id:da016979180284193207] 
\draw  [->,>=stealth]  (326.26,122.17) -- (400.26,122.17) ;

% Text Node
\draw (95.07,92.1) node   [align=left] {(AG)};
% Text Node
\draw (95.07,146.1) node   [align=left] {(AL)};
% Text Node
\draw (94.07,210.1) node   [align=left] {(AD)};
% Text Node
\draw (251.63,119.17) node   [align=center] {(fUAP)\\with high probability};
% Text Node
\draw (472.63,120.67) node   [align=center] {no spurious\\local min in large ball\\$N_{g}<\infty, N_{d} \leq n$\\with high probability};
% Text Node
\draw (292.07,209.1) node   [align=left] {(JKP)};

\draw (180,195) node [anchor=north west][inner sep=0.75pt]   [align=left] {Lem~\ref{lem:infinite small justification}};
\draw (142,102) node [anchor=north west][inner sep=0.75pt]   [align=left] {Lem~\ref{lemma:finite-universal}};
\draw (345,102) node [anchor=north west][inner sep=0.75pt]   [align=left] {Thm~\ref{thm:finiteNd}};

\end{tikzpicture}

\end{center}
 \vspace{-1.2em}
\caption{The diagrammatic summary of the proof in Section \ref{sec:finite}.
We denote \hyperlink{cond:finite_univ}{(Finite universal approximation property)} by (fUAP),
\hyperlink{cond:kernel_point}{(Jacobian kernel point condition)} by (JKP),
and \hyperlink{cond:kernel_ball}{(Jacobian kernel ball condition)} by (JKB).
}
 \vspace{-0.5em}
\label{figure:asumptions2}    
\end{figure*}
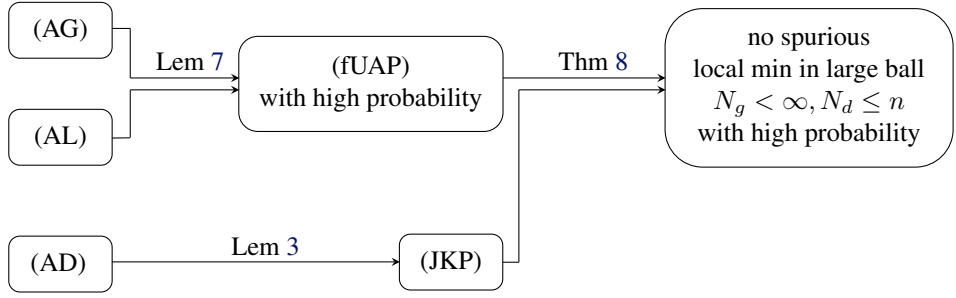

%%%% Large discriminator
%%%%%%%%%%%%%%%%%%%%%%%
\subsection{Large discriminator $(n<N_d<\infty)$}\label{subsec:large discriminator}
Next, consider the case where the discriminator has width $N_d> n$. 
In the small discriminator case, we used the \hyperlink{cond:kernel_point}{(Jacobian kernel point condition)}, which states $\mathrm{rank}(D\Psi(x)^\intercal)=N_d$.
However, this is not possible in the large discriminator case as $\mathrm{rank}(D\Psi(x)^\intercal)\le n<N_d$.
Therefore, we consider the following weaker condition.

\setlength{\leftskip}{0.2in}
\setlength{\rightskip}{0in}

\hypertarget{cond:kernel_ball}{\textbf{(Jacobian kernel ball condition)}}
For any open ball $B\subset \fR^n$,
\[
\bigcap_{x\in B} \ker (D\Psi(x)^\intercal) = \{\bzero{N_d}\}.
\]

\setlength{\leftskip}{0in}
\setlength{\rightskip}{0in}

Since $\nabla_x (\eta^\intercal\Psi(x))=D\Psi(x)^\intercal \eta$,
the \hyperlink{cond:kernel_ball}{(Jacobian kernel ball condition)} implies that
$\eta^\intercal\Psi(x)$ with $\eta\ne 0$ is not a constant function within any open ball $B$, and we can interpret the condition to imply that there is no redundancy in the discriminator feature functions $\psi_1,\dots,\psi_{N_d}$.
The condition holds generically under mild conditions, as characterized by the following lemma.

\begin{lemma}\label{lem:infinite large justification}
Assume $\sigma\colon\fR\rightarrow\fR$ is the sigmoid or the $\tanh$ function.
Then \hyperlink{cond:AD}{(AD)} implies \hyperlink{cond:kernel_ball}{(Jacobian kernel ball condition)} with probability $1$.
\end{lemma}
% Here we provide a brief outline of Lemma~\ref{lem:infinite large justification} while deferring the full proof to the appendix.

\begin{proof}[Proof outline of Lemma~\ref{lem:infinite large justification}]
The random generation of \hyperlink{cond:AD}{(AD)} implies that with probability $1$, all nonzero linear combinations of $\psi_1,\dots,\psi_{N_d}$ are nonconstant, i.e., $f_\eta(x)=\eta^\intercal\Psi(x)$ with $\eta\ne 0$ is not globally constant \citep[Lemma~1]{SUSSMANN1992589}.
Since $\sigma$ is an analytic function, this implies $f_\eta$ with $\eta\ne 0$ is not constant within any open ball $B$.
So $\nabla_x f_\eta(x)=D\Psi(x)^\intercal \eta$ is not identically zero in $B$ and we conclude $\eta\notin \bigcap_{x\in B} \ker (D\Psi(x)^\intercal)$ for any $\eta\ne 0$.
\end{proof}

% Let $X_{\theta} = g_{\theta}(Z)$ be an output random variable from the generator.

We are now ready to state and prove the second main result of this work.

\begin{theorem}\label{thm:infinite_large}
Assume \hyperlink{cond:al}{(AL)}, \hyperlink{cond:AG}{(AG)}, and \hyperlink{cond:AD}{(AD)}.
Then the following statement holds%
\footnote{Once $\psi_j$'s have been generated and the 
\hyperlink{cond:kernel_ball}{(Jacobian kernel ball condition)} holds by Lemma~\ref{lem:infinite large justification}, the conclusion of Theorem~\ref{thm:infinite_large} holds without further probabilistic quantifiers.
}
with probability $1$: for any stationary point $\theta_\mathrm{s}$, if the range of $g_{\theta_\mathrm{s}}(Z)$ contains an open-ball in $\fR^n$, then $J(\theta_\mathrm{s})=0$.
% If a stationary point $\theta_\mathrm{s}$, is such that if the range of $g_{\theta_\mathrm{s}}(Z)$ contains an open-ball in $\fR^n$,
% then $J(\theta_\mathrm{s}) = 0$.
\end{theorem}

\begin{proof}
Following the same steps as in the proof of Theorem~\ref{thm:infinite}, we arrive at \eqref{thm:infinite-key}, which we rewrite as
\[
D\Psi(g_{\theta}(z))^\intercal
r(\theta)
% \begin{bmatrix}
% r_1(\theta)\\
% \vdots\\
% r_{N_d}(\theta)
% \end{bmatrix}
= \bzero{n}.
\]
Thus, for all $z\in\fR^k$,
\[
r(\theta) \in \ker (D\Psi(g_{\theta}(z))^\intercal).
\]
Since the range of $X_{\theta}=g_{\theta}(Z)$ contains an open ball, the \hyperlink{cond:kernel_ball}{(Jacobian kernel ball condition)}, which holds with probability $1$ by Lemma~\ref{lem:infinite large justification}, implies $r(\theta) = \bzero{N_d}$.
\end{proof}

Theorem~\ref{thm:infinite_large} implies that a stationary point may be a spurious stationary point only when the generator's output is degenerate.
One can argue that when $P_X$, the target distribution of $X$, has full-dimensional support, the generator should not converge to a distribution with degenerate support. Indeed, this is what we observe in our experiments of Section~\ref{s:experiments}.

%%%%%%%%%%%%%%%%%%%%%%%%%%%%%%
%%% Finite Width Generator
%%%%%%%%%%%%%%%%%%%%%%%%%%%%%%
\section{Finite-width generator} \label{sec:finite}
Consider a GAN with a two-layer \emph{finite-width} generator $g_\theta\in\mathrm{span} (\{\phi_i\}_{i=1}^{N_g})$ and a two-layer finite-width discriminator $f_{\eta}(x)\in \mathrm{span}(\cD)$.
In this section, we show that $J(\theta)$ has no spurious stationary points within a ball whose radius becomes arbitrarily large (to cover the entire parameter space) as the generator's width $N_g$ goes to infinity.

The finite-width analysis relies on a finite version of the \hyperlink{cond:univ_apx}{(Universal approximation property)} that implies we can approximate a given function as a linear combination of $\{\phi_i\}_{i=1}^{N_g}$.
Let $\delta^{(l)}: \fR^k \rightarrow \fR^n$ have the delta function on the $l$\nobreakdash-th component
and zero functions for all other components, i.e.,
\[
\left[\delta^{(l)}(z)\right]_i
=\begin{cases}
0 &\mbox{ if $i\neq l$}\\
\delta(z) &\mbox{ if $i=l$}
\end{cases}
\]
 for $1\leq l\leq n$.
% Although the delta ``function'' is not truly a function but rather a measure,
% this distinction is not needed as we only use the delta function as a notational shorthand. 
% \noa{Do we want to make an additional comment regrading this?}
% \ryu{Let's just remove this sentence entirely and avoid the issue.}
% \noa{cool}

\setlength{\leftskip}{0.2in}
\setlength{\rightskip}{0in}
\hypertarget{cond:finite_univ}{\textbf{(Finite universal approximation property)}}
For a given $\varepsilon>0$, there exists a large enough $N_g\in \mathbb{N}$ and  $\phi_1, \ldots, \phi_{N_g}\in\cG$ such that
there exists $\{\theta_i^{(\varepsilon,l)}\in \fR\,|\, 1\leq i\leq N_g, 1\leq l\leq n\}$ satisfying
\begin{align*}
    &\left|
\EZ{ \left(\sum_{i=1}^{N_g} \theta_i^{(\varepsilon,l)}\phi_i(Z) - \delta^{(l)}(Z)\right)^\intercal \!\!\!f(Z)}
\right|\\
&\qquad\qquad\qquad\qquad\qquad< \varepsilon\sup_{z\in\fR^k}\{\|f(z)\|_2+\|Df(z)\|\}
\end{align*}
for all coordinates $l=1,\dots,n$, and for any continuously differentiable $f\colon \fR^k\rightarrow\fR^n$ such that $\sup_{z\in\fR^k}\{\|f(z)\|_2+\|Df(z)\|\}<\infty$.

\setlength{\leftskip}{0in}
\setlength{\rightskip}{0in}

This \hyperlink{cond:finite_univ}{(Finite universal approximation property)} holds with high probability when the width $N_g$ is sufficiently large and the weights and biases of the generator feature functions $\phi_1,\dots\phi_{N_g}$ are randomly generated.

\begin{lemma}
\label{lemma:finite-universal}
% \ryu{XXX Taeho please check. I think this lemma needs to assume (AL)
% Dependence through Lemma 11
% XXX}
Assume \hyperlink{cond:al}{(AL)} and \hyperlink{cond:AG}{(AG)}.
Assume the first $n$ parameters  $\{\kappa_i\}_{i=1}^n$ are chosen so that $\{\phi_i\}_{i=1}^n$ are constant functions spanning the sample space $\fR^n$.
Assume the remaining parameters $\{\kappa_i\}_{i=n+1}^{N_g}$ are sampled (IID) from a probability distribution that has a continuous and strictly positive density function.
Then for any $\varepsilon>0$ and $\zeta>0$, there exists large enough
\footnote{A quantitative bound on $N_g$ can be established with careful bookkeeping. 
Specifically, using  \eqref{eqn:integral_to_finite_sum_Hilbert} of Appendix~\ref{app:proof of finite-universal}, we can quantify $N_g$ as a function of $C_K$.
This $C_K$, which serves a similar role as the $C$ of \citep[Theorem~1]{rahimi2007random}, can also be quantified as a function of $\epsilon$.
However, the resulting bound is complicated and loose.
} 
$N_g$ such that \hyperlink{cond:finite_univ}{(Finite universal approximation property)} with $\varepsilon$ holds with probability at least $1-\zeta$.
\end{lemma}

Remember that the parameters define the generator feature functions through $\phi_i(x)=\phi_i(x;\kappa_i)$ for $1\le i\le N_g$.
By choosing the first $n$ parameters in this way, we are effectively providing a trainable bias term in the output layer of the generator. Note that most universal approximation results consider the approximation of functions, while \hyperlink{cond:finite_univ}{(Finite universal approximation property)} requires the approximation of the delta function, which is not truly a function.

% Here we provide a brief outline of Lemma~\ref{lemma:finite-universal} while deferring the full proof to the appendix.

\begin{figure*}[ht]
% \vspace{-5mm}
\centering
\begin{tabular}{ccc}
% \hspace{-7mm}
% \hspace{-9mm}
% \subfigure[Convergence of the loss function $L$]{
%       \includegraphics[width=0.45\textwidth]{fig2}}
% \end{tabular}\\
% \vspace{-7mm}
\hspace{-4mm}
\subfigure [Samples from true distribution $P_X$]{
      \raisebox{0.1 \height} {
      \includegraphics[width=0.3\textwidth]{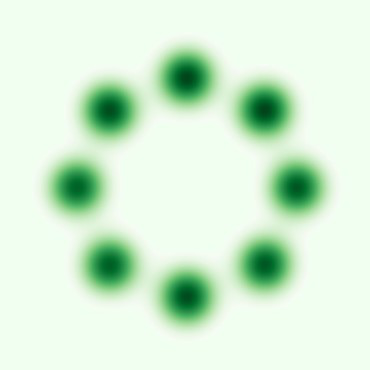}
      }
      }
&
\hspace{-6mm}
\subfigure[Samples from generator $g_\theta(Z)$]{
      \raisebox{0.1 \height} {
      \includegraphics[width=0.3\textwidth]{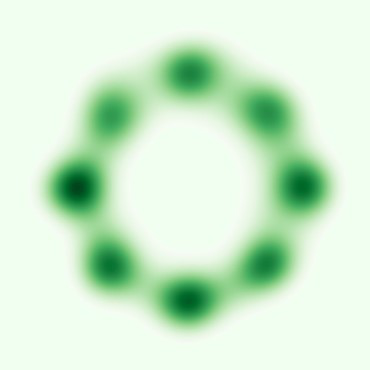}
      }}
&
\hspace{-6mm}
\subfigure[Convergence of the loss functions $J$ and $L$]{
      \raisebox{0.05 \height} {
      \includegraphics[width=0.38\textwidth]{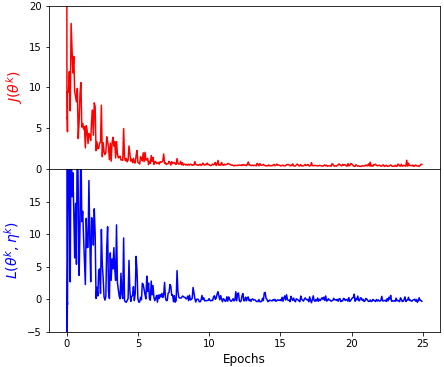}
      }
      }
\end{tabular}
 \vspace{-1.2em}
\caption{Samples and loss functions with a mixture of 8 Gaussians, $X\in \fR^2$, $Z\in \fR^2$, $N_g=5,000$, and $N_d=1,000$.
The generator accurately learns the sampling distribution, and the loss functions converge to $0$. The code is available at \url{https://github.com/sehyunkwon/Infinite-WGAN}.}
\vspace{-0.5em}
\label{fig:samples}
\end{figure*}

%%% Assumptions
%%%%%%%%%%%%%%%
\begin{proof}[Proof outline of Lemma~\ref{lemma:finite-universal}]
Here, we illustrate the proof in the case of $n=1$. The general $n\ge 1$ case requires similar reasoning but more complicated notation.

First, we define the smooth approximation of $\delta$ by
\[
\tilde{\delta}^\varepsilon(z)=
\frac{C}{\varepsilon^k}e^{-\|z/\varepsilon\|^2_2},
\]
where $C$ is a constant (depending on $k$ but not $\varepsilon$) such that 
\[ 
\int_{\fR^k}\tilde{\delta}^\varepsilon(z)\,dz=1.
\]
We argue that $\tilde{\delta}^\varepsilon(z)\approx \delta$ in the sense made precise in Lemma~\ref{lemma:w11-delta} of the appendix.

Next, we approximate $\tilde{\delta}^\varepsilon$ with the random feature functions.
Using the arguments of \citep[Theorem~2]{barron1993universal} and \citep[Section~4.2]{Telgarsky2020}, we show that there exists a bounded density $m(\kappa)$ and $\kappa_1 \in \fR^{k+1}$ such that $\phi_1 = \phi(z;\kappa_1)$ is a nonzero constant function and
\[
\tilde{\delta}^\varepsilon(z)\approx 
\theta^\varepsilon_1 \phi_1(z)
% \tilde{\delta}^\varepsilon(0) 
+ \int\phi(z;\kappa)\,m(\kappa) \, d\kappa
\]
for some $\theta^\varepsilon_1 \in \fR$.
For large $K>0$,
\[
\int\phi(z;\kappa)\,m(\kappa)d\kappa\approx \int\phi(z;\kappa)\,m(\kappa)\mathbf{1}_{\{\|\kappa\|\le K\}}(\kappa)\, d\kappa,
\]
where $\mathbf{1}_{\{\|\kappa\|\le K\}}$ is the 0-1 indicator function.
Write $p(\kappa)$ for the continuous and strictly positive density function of the distribution generating $\kappa$. Then $\sup_{\kappa} \{m(\kappa)\mathbf{1}_{\{\|\kappa\|\le K\}}(\kappa)/p(\kappa)\}<\infty$,
and this allows us to use random feature learning arguments of \citep{kitchensink2008}.
By \citep[Lemma~1]{kitchensink2008}, there exists a large enough $N_g$ such that there exist weights $\{\theta^\varepsilon_i\}_{i=2}^{N_g}$ such that
\[
\sum^{N_g}_{i=2}\theta^\varepsilon_{i}\phi(z;\kappa_i)
\approx
\int\phi(z;\kappa)\,m(\kappa)\mathbf{1}_{\{\|\kappa\|\le K\}}(\kappa)\, d\kappa
\]
with probability $1-\zeta$.
Finally, we complete the proof by chaining the $\approx$ steps.
\end{proof}

We are now ready to state and prove the third main result of this work:
our GAN with an finite-width generator and small discriminator has no spurious stationary points within a large ball around the origin.
\begin{theorem}\label{thm:finiteNd}
Let $N_d\leq n$.
Assume \hyperlink{cond:al}{(AL)}, \hyperlink{cond:AG}{(AG)}, and \hyperlink{cond:AD}{(AD)}.
Assume the generator feature functions are generated randomly as in Lemma~\ref{lemma:finite-universal}.
For any $C_{}>0$ and $\zeta>0$, there exists a large enough $N_g\in \mathbb{N}$ such that the following statement holds with probability at least $1-\zeta$: any stationary point $\theta_\mathrm{s}\in \fR^{N_g}$ satisfying $\|\theta_\mathrm{s}\|_1\le C_{}$ is a global minimum.\footnote{The randomness comes from the random generation of $\psi_j$'s and $\phi_i$'s described in \hyperlink{cond:AD}{(AD)} and Lemma~\ref{lemma:finite-universal}.
Once the \hyperlink{cond:kernel_point}{(Jacobian kernel point condition)} and \hyperlink{cond:finite_univ}{(Finite universal approximation property)} holds
(an event with probability at least $1-\zeta$) the conclusion of Theorem~\ref{thm:finiteNd} holds without further probabilistic quantifiers.
Again, the size of $N_g$ can be quantified through careful bookkeeping.
}
\end{theorem}

\begin{proof}[Proof of Theorem \ref{thm:finiteNd}]
Since $\phi$ is bounded by \hyperlink{cond:AG}{(AG)} and $\|\theta_\mathrm{s}\|_1$ is bounded, the output of $g_{\theta_\mathrm{s}}$ is also bounded, and
\[
\sup_{\|\theta_\mathrm{s}\|_1\leq C} \|g_{\theta_\mathrm{s}}(\bzero{n})\|_2<\infty.
\]
The \hyperlink{cond:kernel_point}{(Jacobian kernel point condition)}, which holds with probability $1$ by Lemma~\ref{lem:infinite small justification}, implies
\[
C_1 \stackrel{\Delta}{=} \inf_{\|x\|_2\leq  \left(\underset{{\|\theta_\mathrm{s}\|_1\leq C}}{\sup} \|g_{\theta_\mathrm{s}}(\bzero{n})\|_2\right)}\tau_\mathrm{min}(D\Psi(x)^\intercal)>0,
\]
where $\tau_\mathrm{min}$ denotes the $N_d$-th singular value. We use the fact that $\tau_\mathrm{min}(D\Psi(x)^\intercal)$ is a continuous function of $x$ and the infimum over a compact set of a continuous positive function is positive.
(We use $\tau_\mathrm{min}$ to denote the minimum singular value,
rather than the standard $\sigma_\mathrm{min}$ to avoid confusion with the $\sigma$ denoting the activation function.)
By \hyperlink{cond:AD}{(AD)},
\[
C_2 \stackrel{\Delta}{=} \max_{j=1,\dots,N_d}\sup_{x\in\fR^n}\{\|\nabla \psi_j(x)\|+\|\nabla^2 \psi_j(x)\|\}\in (0,\infty)
\]
By Lemma~\ref{lemma:finite-universal}, there exists a large enough $N_g$ such that \hyperlink{cond:finite_univ}{(Finite universal approximation property)} with
\[
\varepsilon = \frac{C_1 q_Z(\bzero{k})}{2C_2N_dn}
\]
holds with probability $1-\zeta$.

Let $ \theta_\mathrm{s}$ be a stationary point satisfying $ \|\theta_\mathrm{s}\|_1\le C$.
However, assume for contradiction that $J(\theta_\mathrm{s})\ne 0$, i.e., $r(\theta_\mathrm{s})\ne 0$.
Then
\begin{align*}
\frac{\partial}{\partial \theta_i}J(\theta_\mathrm{s})=\EZ{\sum_{j=1}^{N_d} r_j(\theta_\mathrm{s}) \nabla\psi_j(g_{\theta_\mathrm{s}}(Z))^\intercal \phi_i(Z)}=0 
\end{align*}
for all $1\leq i \leq N_g$.
Define the normalized residual vector $\rh=(1/\|r\|_2)r$, and write
% i.e.,
% \[
% \rh_j(\theta)=\frac{1}{\|r(\theta)\|_2}r_j(\theta)
% \]
% for $1\leq j\leq N_d$.
\begin{equation}
\EZ{\sum_{j=1}^{N_d} \rh_j(\theta) \nabla\psi_j(g_{\theta}(Z))^\intercal \phi_i(Z)} = 0
\label{eq:finite-zero-phi}
\end{equation}
for all $1\leq i \leq N_g$.

Now consider
\begin{align*}
&\left|\sum_{j=1}^{N_d}\rh_j(\theta)\frac{\partial}{\partial x_l}\psi_j(g_{\theta}(\bzero{n})) q_Z(\bzero{n})\right|\\
&=
\Bigg|\mathbb{E}_Z\Bigg[
\underbrace{\sum_{j=1}^{N_d} \rh_j(\theta)\nabla\psi_j(g_{\theta}(Z))^\intercal}_{\stackrel{\Delta}{=}f(Z)^\intercal} \delta^{(l)} (Z)\Bigg]\Bigg|\\
&=
\left|\EZ{f(Z)^\intercal
\left(\delta^{(l)} (Z)-\sum_{i=1}^{N_g} \theta_i^{(l, \varepsilon)}\phi_i(z)\right)}\right|\\
&< 
\varepsilon\sup_{z\in \fR^k}\{\|f(z)\|+\|Df(z)\|\}
\\
&\le \varepsilon\sum_{j=1}^{N_d} |\rh_j(\theta)| \sup_{x\in \fR^n}\{\|\nabla \psi_j(x)\|+ \|\nabla^2\psi_j(x)\|  \}\\
&\le\varepsilon C_2 N_d,
\end{align*}
where the first equality follows from the definition of $\delta^{(l)}$,
the second equality follows from \eqref{eq:finite-zero-phi},
the first inequality follows from the the \hyperlink{cond:finite_univ}{(Finite universal approximation property)}, 
the second inequality follows from the triangle inequality of the norm,
and the third inequality follows from the definition of $C_2$ and the fact that the normalized residual satisfies $|\rh_j(\theta)|\le 1$ for all $j$.
By summing this result over $1\le l\le n$ and using the bound $\|\cdot\|_2\le \|\cdot\|_1$, we get
% \[
% \left|\sum_{j=1}^{N_d}\rh_j(\theta)\frac{\partial}{\partial x_l}\psi_j(g_{\theta}(\bzero{n})) q_Z(\bzero{n}) \right|
% <(C_2 N_d)\cdot\varepsilon,
% \]
% which implies
\vspace{-0.05in}
\[
\left\|D\Psi(g_{\theta}(\bzero{n}))^\intercal \rh(\theta)\right\|_2
<\varepsilon\frac{C_2 N_dn}{q_Z(\bzero{n})}.
\]
\vspace{-0.05in}
Finally, we arrive at
\begin{align*}
C_1= C_1  \|\rh(\theta)\|_2&\le \left\|D\Psi(g_{\theta}(\bzero{n}))^\intercal \rh(\theta)\right\|_2\\
&< 
\varepsilon\frac{C_2 N_dn}{q_Z(\bzero{k})}
=\frac{C_1}{2},
\end{align*}
\vspace{-0.1in}
which is a contradiction.
\end{proof}

\begin{figure*}[ht]
% \vspace{-5mm}
\centering
\begin{tabular}{cc}
\hspace{-7mm}
\subfigure[Loss landscape with $N_g=2$]{
      {
      \raisebox{0 \height} {
      \includegraphics[width=0.53\textwidth]{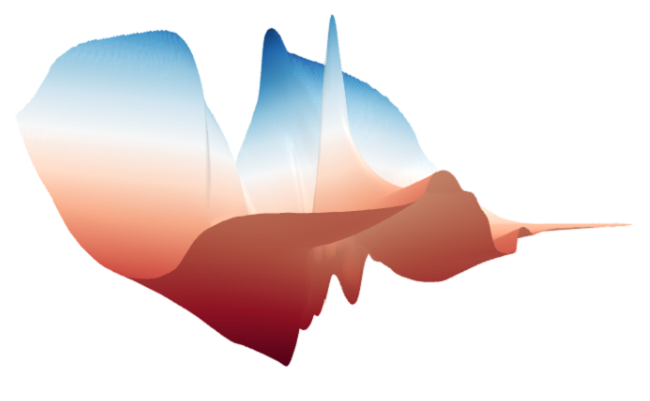}
      }
      }
      }
% &
% % \hspace{-9mm}
% \subfigure[Landscape with $N_g=2$.]{
%       \includegraphics[width=0.45\textwidth]{fig6}}
% \end{tabular}\\
% % \vspace{-7mm}
% \begin{tabular}{cc}
% \subfigure[Landscape with $N_g=10$.]{
%       \includegraphics[width=0.4\textwidth]{fig5}}
% &
\hspace{-17mm}
&
% \!\!\!\!\!\!
\subfigure[Loss landscape with $N_g=10$]{
      {
      \raisebox{0.1 \height} {
      \includegraphics[width=0.53\textwidth]{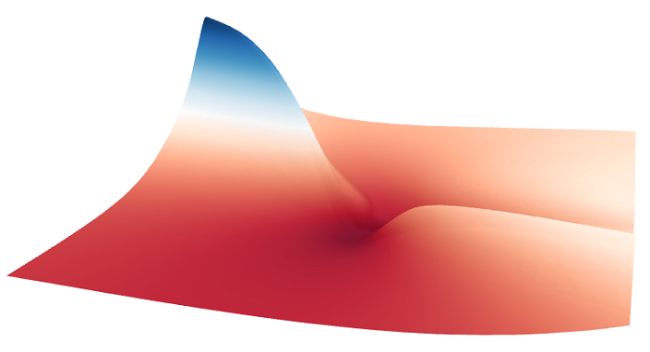}
      }
      }
      }
\end{tabular}
\vspace{0.0em}
\caption{
The landscape of $J(\theta)$ for a mixture of two Gaussians with generator widths $N_g=2$ and $N_g=10$.
The first $N_g=2$ example has multiple non-global local minima.
The second $N_g=10$ example has no spurious stationary points despite the landscape being clearly non-convex.
We provide corresponding contour plots in the appendix.
}
\vspace{-0.5em}
\label{fig:landscape}
\end{figure*}

%%%%%%%%%%%%%%%%%%%%%%%%%%%%%%%%%%%%%%%%%%%%%%%%%%
%%% Infinite Width Generator
%%%%%%%%%%%%%%%%%%%%%%%%%%%%%%%%%%%%%%%%%%%%%%%%%%
\section{Experiments} \label{s:experiments}
Figure~\ref{fig:samples} presents an experiment with a mixture of $8$ Gaussians and $N_g=5,000$. The experiments demonstrate the sufficiency of two-layer networks with random features and that the training does not encounter local minima when $N_g$ is large.

Figure~\ref{fig:landscape} visualizes the loss landscape with generator widths $N_g=2$ and $N_g=10$. For the $N_g=10$ case, the parameter space was projected down to a 2D space defined by random directions, as recommended by \citet{li2018visualizing}. We observe the landscape becomes more favorable with larger width.

% \noa{
% The estimated expected loss converges to zero as described in
% Figure \ref{fig:samples} where we both plot $J(\theta^t)$ and $L(\theta^t, \eta^t)$
% at every XXX 100 XXX batches.
% For more details, please check Appendix \ref{app:}.
% }

\section{Conclusion}
In this work, we presented an infinite-width analysis of a WGAN and established that no spurious stationary points exist under certain conditions.

At the same time, however, we point out that the infinite-width analysis does simplify away (hide) some finite phenomena. One such issue we encountered in our experiments was nearly vanishing gradients, which can occur despite the absence of spurious stationary points. A quantitative finite-width analysis establishing explicit bounds may provide an understanding and remedies to such issues and, therefore, is an interesting direction of future work.

\section*{Acknowledgements}
AN was supported by Basic Science Research Program through the National Research Foundation of Korea (NRF)
funded by the Ministry of Education (2021R1F1A105956711).
TY, SK, and EKR were supported by the National Research Foundation of Korea (NRF) Grant funded by the Korean Government (MSIP) [No. 2020R1F1A1A01072877], the National Research Foundation of Korea (NRF) Grant funded by the Korean Government (MSIP) [No. 2017R1A5A1015626], by the New Faculty Startup Fund from Seoul National University, and by the AI Institute of Seoul National University (AIIS) through its AI Frontier Research Grant (No. 0670-20200015) in 2020.
We thank Jisun Park for reviewing the manuscript and providing valuable feedback. 
Finally, we thank the anonymous referees for their suggestions on improving the clarity of the exposition and including the diagrammatic summaries of Figures~\ref{figure:asumptions1} and \ref{figure:asumptions2}.

%%%%%%%%%%%%%%%%%%%%%%%%%%%%%%%%%%%%%%%%%%%%%%%%%%
%%% Bibliography
%%%%%%%%%%%%%%%%%%%%%%%%%%%%%%%%%%%%%%%%%%%%%%%%%%
\bibliography{gan}
\bibliographystyle{icml2021}


\onecolumn
\newpage
\appendix

\section{Omitted proofs}
\subsection{Proof of Lemma~\ref{lem:universal}}
% The precise statement of Theorem~1 of \citep{HORNIK1991251} is as follows
    \begin{theorem}[\protect{\citep[Theorem~1]{HORNIK1991251}}]\label{thm:finite_universal}
Let $\sigma\colon\fR\rightarrow\fR$ be bounded and nonconstant and $P\in \mathcal{M}(\fR^k)$ be a finite measure.
Then for any $f\in L^1(P)$ and $\varepsilon>0$,
there exists $N\in \mathbb{N}$ and $\{(\theta_i,a_i,b_i)\in \fR\times \fR^k\times \fR\}_{i=1}^N$ such that 
\[
\int_{\fR^k}\left|\sum_{i=1}^N\theta_i\sigma (a_i^\intercal z-b_i)-f(z)\right|\,dP(z)
<\varepsilon.
\]
\end{theorem}
To clarify, $f\colon \fR^k\rightarrow \fR$ in \citep[Theorem~1]{HORNIK1991251}.

\begin{proof}[Proof of Lemma~\ref{lem:universal}]
Let $f\colon\fR^k\rightarrow\fR^n$ such that $\EZ{\|f(Z)\|_2}<\infty$.
By \hyperlink{cond:AG}{(AG)}, $\sigma_\mathrm{g}$ is a bounded nonconstant function.
For $l=1,\dots,n$, Theorem~\ref{thm:finite_universal} provides us with $N_l\in \mathbb{N}$ and $\{(\theta_i^{(l)},a_i^{(l)},b_i^{(l)})\in \fR\times \fR^k\times \fR\}_{i=1}^{N_l}$
such that
\[
h_l(z)=\sum_{i=1}^{N_l}\theta_i^{(l)}\sigma_\mathrm{g} ((a_i^{(l)})^\intercal z-b_i^{(l)})
\]
satisfies
\begin{equation}
\int_{\fR^k}\left|h_l(z)-f_l(z)\right|\,q_Z(z)dz
<\frac{\varepsilon}{2n},
\label{eq:hornikapplication}
\end{equation}
where $f_l(z)$ is the $l$\nobreakdash-th coordinate of $f(z)\in\fR^n$ for $l = 1,\dots,n$.
Let $\ell_g = \lim_{r\to -\infty} \sigma_g (r)$.
Let
\[
A^{(l)}_i=
\begin{bmatrix}
0\\\vdots\\0 \\(a_i^{(l)})^\intercal\\0\\\vdots\\0
\end{bmatrix}
\leftarrow\text{$l$\nobreakdash-th row}
,\qquad
\mathbf{b}^{(l),r}_i=
\begin{bmatrix}
r\\\vdots\\r\\
b_i^{(l)}\\r\\\vdots\\r
\end{bmatrix},
\qquad
e_{-l} =
\begin{bmatrix}
1\\\vdots\\1\\
0\\1\\\vdots\\1
\end{bmatrix}
\leftarrow\text{on $l$\nobreakdash-th coordinates}
\]
and
\[
\tilde{f}^{(l),r}(z)=
-\left( \sum_{i=1}^{N_l} \theta_i^{(l)} \right) \ell_g e_{-l}+
\sum_{i=1}^{N_l}\theta_i^{(l)}\sigma_\mathrm{g} (A_i^{(l)} z-\mathbf{b}^{(l),r}_i).
\]
% and define $e_{-l}\in \fR^N$ as 
% i.e., $e_{-l}$ is the vector with all $1$'s except in the $l$\nobreakdash-th coordinate which has a $0$.
Then, for each $l=1,\dots,n$, we have
$\tilde{f}^{(l),r}_j = \sum_{i=1}^{N_l} \theta_i^{(l)} (\sigma_g(-r) - \ell_g) \to 0$ as $r \to \infty$ if $j\ne l$, while
$\tilde{f}^{(l),r}_l = h_l(z)$.
Because $\sigma_g$ is bounded, by Lebesgue's dominated convergence theorem, we obtain
\[
\lim_{r\rightarrow \infty}
\int_{\fR^k}
\left\|
\begin{bmatrix}
h_1(z)\\
\vdots\\
h_n(z)
\end{bmatrix}
-\sum^n_{l=1}\tilde{f}^{(l),r}(z)
\right\|_1\,q_Z(z)dz = 0.
\]
% by Lebesgue's dominated convergence theorem.
Therefore, there exists a large enough $r_\mathrm{big}>0$ such that
\[
\int_{\fR^k}
\left\|
\begin{bmatrix}
h_1(z)\\
\vdots\\
h_n(z)
\end{bmatrix}
-\sum^n_{l=1}\tilde{f}^{(l),r_\mathrm{big}}(z)
\right\|_1\,q_Z(z)dz<\frac{\varepsilon}{2}
\]
and we conclude with \eqref{eq:hornikapplication} that
\[
\int_{\fR^k}
\left\|
f(z)
-\sum^n_{l=1}\tilde{f}^{(l),r_\mathrm{big}}(z)
\right\|_1\,q_Z(z)dz<\varepsilon.
\]
Note that
\[
\sum^n_{l=1}\tilde{f}^{(l),r_\mathrm{big}}(z)\in \overline{\mathrm{span}} (\cG).
\]
Therefore, using the bound $\|\cdot\|_2\le  \|\cdot\|_1$, we get
\[
\int_{\fR^k}
\left\|
f(z)
-\sum^n_{l=1}\tilde{f}^{(l),r_\mathrm{big}}(z)
\right\|_2\,q_Z(z)dz<\varepsilon.
\]
\end{proof}

\subsection{Proof of Lemma~\ref{lem:dual-universal}}
\begin{proof}[Proof of Lemma~\ref{lem:dual-universal}]
% XXX standard dual characterization in functional analysis. XXX
Because $h$ is bounded and $q_Z(z)\,dz$ is a probability measure, we have $\EZ{\|h(Z)\|_2} < \infty$.
Therefore, for any $\varepsilon > 0$, there exists $\theta_\varepsilon$ such that $\EZ{\left\| g_{\theta_\varepsilon}(Z) - h(Z) \right\|_2} < \varepsilon$.
Observe that
\begin{align*}
    \EZ {g_{\theta_\varepsilon}^\intercal(Z) h(Z)}
    % = \mathbb{E}_Z \left[ \int h^\intercal(Z) \phi(Z;\kappa) \, d\theta_\varepsilon(\kappa) \right]
    & = \int_{\fR^k} \int_{\fR^p} h^\intercal(z) \phi(z;\kappa) \, d\theta_\varepsilon(\kappa) \, q_Z(z)\,dz \\
    & = \int_{\fR^p} \int_{\fR^k} h^\intercal(z) \phi(z;\kappa) q_Z(z)\,dz \,  d\theta_\varepsilon(\kappa) \\
    & = \int \EZ{h^\intercal(Z)\phi(Z;\kappa)} \, d\theta_\varepsilon(\kappa) = 0.
\end{align*}
Here the change in the order of integration is valid because $\phi(z;\kappa) = \sigma_\mathrm{g} (\kappa_w z + \kappa_b) \le \|\sigma_\mathrm{g}\|_\infty$ and the total mass of $\theta_\varepsilon$ is finite, so that
\begin{align*}
    \int_{\fR^k} \int_{\fR^p} \left\| h^\intercal(z) \phi(z;\kappa) \right\|_2 \, d\theta_\varepsilon(\kappa) \, q_Z(z)\,dz
    \le n\|h\|_{\infty} \left\|\sigma_\mathrm{g}\right\|_{\infty} \theta_\varepsilon\left(\fR^p\right) < \infty.
\end{align*}
To clarify, the $\|\cdot\|_\infty$ for $\|\sigma_\mathrm{g}\|_\infty $ is the standard supremum norm for $L^\infty$ spaces
while $\|h\|_{\infty}=\max_{1\le i\le n}\|h_i\|_\infty$ where $h_i(z)$ is the $i$\nobreakdash-th coordinate of $h(z)\in \fR^n$.
Finally, we have
\begin{align*}
    \EZ{\|h(Z)\|_2^2} = \EZ{h^\intercal(Z) \left(h(Z) - g_{\theta_\varepsilon}(Z)\right)} &\le \|h\|_\infty \EZ{\left\|h(Z) - g_{\theta_\varepsilon}(Z)\right\|_1} \\
    & \le \|h\|_\infty \EZ{\sqrt{n} \left\|h(Z) - g_{\theta_\varepsilon}(Z)\right\|_2} < \varepsilon \sqrt{n} \|h\|_\infty.
\end{align*}
To clarify, $\left\|h(Z) - g_{\theta_\varepsilon}(Z)\right\|_1$ denotes the $\ell^1$ norm on the vector in $\fR^n$ for each $z$.
Now by letting $\varepsilon\rightarrow 0$, we have
\begin{align*}
    0 = \EZ{\|h(Z)\|_2^2} = \int \|h(z)\|_2^2 \, q_Z(z)\,dz .
\end{align*}
Since $q_Z$ is continuous and positive everywhere, we conclude that $h(z) = 0$ for all $z \in \fR^k$.
\end{proof}

\subsection{Proof of Lemma~\ref{lem:infinite large justification}}
%%%%%%%%%%%%%%%%%%%%%%%%%%%%%%%%%%%%%%%%%%%%%%%%%%%%%%%%%%%%%%%%%%%%%%%%%%%%%%%
%%%%%%%%%%%%%%%%%%%%%%%%%%%%%%%%%%%%%%%%%%%%%%%%%%%%%%%%%%%%%%%%%%%%%%%%%%%%%%%

% We rely on the following result in the proof of Lemma~\ref{lem:infinite large justification}.
\begin{theorem}[\protect{\citep[Lemma~1]{SUSSMANN1992589}}]
\label{thm:sussman}
Let $\sigma=\tanh$.
Assume
\[
C_0+
\sum^{N}_{j=1}\eta_j\sigma (a_j^\intercal x+b_j)=
C
\]
for all $x\in \fR^n$, where $\eta_j\ne 0$ and $a_j\ne 0$ for $1\le j\le N$.
If there exists no distinct indices $i$ and $j$ such that $(a_i,b_i)=\pm (a_j,b_j)$,
then $N=0$ (the sum vanishes) and $C_0=C$.
\end{theorem}

\begin{proof}[Proof of Lemma~\ref{lem:infinite large justification}]
First consider the case where $\sigma=\tanh$.
With probability $1$, the condition of Theorem~\ref{thm:sussman} holds, and
\begin{align*}
    F(x) \stackrel{\Delta}{=} \sum_{j=1}^{N_d} \eta_j \psi_j (x)
\end{align*}
with $\eta\ne 0$ is not constant.
% i.e., $f_\eta(x)=\eta^\intercal\Psi(x)$ with $\eta\ne 0$ 
% XXX Assuming a priori that generic linear combination of $\psi_j$'s cannot be a globally constant function, \hyperlink{cond:kernel_ball}{(Jacobian kernel ball condition)} holds true. XXX
Since $\sigma\colon\fR\rightarrow\fR$ is analytic on $\fR$, it has a power series expansion
\begin{align*}
    \sigma (t) = \sum_{\nu=0}^\infty s_{\nu} \, t^\nu.
\end{align*}
Suppose that $0 \ne \eta \in \bigcap_{x\in B} \ker (D\Psi(x)^\intercal)$. Then 
\[
\sum_{j=1}^{N_d} \eta_j \nabla \psi_j (x) \equiv 0
\]
for $x\in B$, and
\begin{align*}
    F(x) \stackrel{\Delta}{=} \sum_{j=1}^{N_d} \eta_j \psi_j (x) = \sum_{j=1}^{N_d} \eta_j \sum_{\nu=0}^\infty s_{\nu} (a_j^\intercal x + b_j)^\nu
\end{align*}
is constant for $x\in B$.
Fix any $x_0 \in B$ and $u \in \fR^n$.
Let $\alpha_j = a_j^\intercal u$ and $\beta_j = a_j^\intercal x_0 + b_j$.
Then for $x_u(t)\stackrel{\Delta}{=}x_0+tu$,
\begin{align*}
    F(x_u(t)) &= \sum_{j=1}^{N_d} \eta_j \sum_{\nu=0}^\infty s_{\nu} (t\alpha_j + \beta_j)^\nu \\
    &= \sum_{j=1}^{N_d} \eta_j \sum_{\nu=0}^\infty s_{\nu} \sum_{m=0}^\nu \binom{\nu}{m} (\alpha_j t)^m \beta_j^{\nu-m} \\
    &= \sum_{m=0}^\infty \left( \sum_{j=1}^{N_d} \sum_{\nu \ge m} \eta_j s_{\nu} \binom{\nu}{m} \alpha_j^m \beta_j^{\nu-m} \right) t^m \\
    &\stackrel{\Delta}{=} F_0 + \sum_{m=1}^\infty F_m t^m
\end{align*}
is constant within $t \in (-\varepsilon,\varepsilon)$ for some $\varepsilon > 0$.
(Order of summations can be freely interchanged because power series for $\sigma$ are absolutely convergent for any choice of $t$.)
But then $F_m$ must be zero for all $m \ge 1$, since $0 = \frac{d^m}{dt^m} \sum_{j=1}^{N_d} \eta_j \psi_j (x_u(0)) = m!\,F_m$.
Therefore, in fact, $F(x_u(t)) \equiv F_0$ for all $t\in \fR$, and $F_0 = F(x_0)$ does not depend on $u$.
That is, $F$ is a constant function on $\fR^n$.
This implies that $\eta = \bzero{N_d}$, which contradicts the assumption $\eta \ne 0$.

We extend the conclusion to the sigmoid function by noting that
\[
\frac{1}{1+e^{-r}}=\frac{\tanh(r/2)+1}{2},
\]
i.e., the sigmoid function is obtained by scaling the input of $\tanh$, adding a constant, and scaling the output.
\end{proof}
% Since $r(\theta)\in\ker(D\Psi(\tx)^\intercal)$ implies that 
% \[
% \mbox{diag}(\sigma_1'(a_1^\intercal\tx+b_1), \ldots, \sigma_{N_d}'(a_{N_d}^\intercal\tx+b_{N_d}))\cdot r(\theta) \in \ker (A)
% \]
% where $A = \begin{bmatrix}a_1& a_2& \cdots& a_{N_d}\end{bmatrix}$.

\subsection{Proof of Lemma~\ref{lemma:finite-universal}}
\label{app:proof of finite-universal}
Recall that we defined
\[
\tilde{\delta}^\varepsilon(z) = \frac{\pi^{-k/2}}{\varepsilon^k} e^{-\|z/\varepsilon\|_2^2},
\]
so that $\int_{\fR^k} \tilde{\delta}^\varepsilon(z) \, dz = 1$ for all $\varepsilon > 0$.

\begin{lemma}
\label{lemma:w11-delta}
Assume \hyperlink{cond:al}{(AL)}.
There exists a constant $C_\delta$ depending only on $k$ but not on $\varepsilon>0$ such that 
\[
\left| \mathbb{E}_Z \left[ \left( \tilde{\delta}^\varepsilon(Z) - \delta(Z) \right) f(Z) \right] \right|
<
C_\delta \, \varepsilon \sup_{z\in \fR^k} \left( |f(z)| + \|\nabla f(z)\| \right)
\]
for all differentiable $f\colon\fR^k\rightarrow\fR$ such that $\sup_{z\in \fR^k} \left( |f(z)| + \|\nabla f(z)\| \right) <\infty$.
Here $\|\cdot\|$ denotes the operator norm, which coincides with the vector $\ell^2$ norm on $\fR^k$.
\end{lemma}

\begin{proof}
%[Proof of Lemma~\ref{lemma:w11-delta}]
Let $M=\|f\|_\infty$, $L_f = \sup_{z\in \fR^k} \|\nabla f(z)\|$ and let $L_Z$ be the Lipschitz constant of $q_Z(z)$.
Then for any $z\in \fR^k$,
\[
|f(z) q_Z(z) - f(0) q_Z(0)| \le |f(z)| |q_Z(z)-q_z(0)| + |f(z)-f(0)| q_Z(0) \le M L_Z \|z\| + L_f \|z\| q_Z(0).
\]
Integrating both sides over $z\in\fR^k$ with respect to $\tilde{\delta}^\varepsilon(z) \, dz$ gives
\begin{align*}
% \label{eqn:fqz_integral_bound}
\int_{\fR^k} |f(z)q_Z(z) - f(0)q_Z(0)| \, \tilde{\delta}^\varepsilon(z) \, dz & \le \int_{\fR^k} (ML_Z + L_f q_Z(0)) \|z\| \tilde{\delta}^\varepsilon (z) \, dz \\
& = \int_{\fR^k} (ML_Z + L_f q_Z(0)) \frac{\pi^{-k/2}}{\varepsilon^k} \|z\|e^{-\|z/\varepsilon\|_2^2} \,dz.
\end{align*}
Using change of variables, we rewrite and bound the last integral as
\begin{align*}
(ML_Z + L_f q_Z(0)) \pi^{-k/2} \varepsilon \int_{\fR^k} \|z\| e^{-\|z\|_2^2} \, dz &\le
\max\{L_Z, q_Z(0)\} \pi^{-k/2} \left( \int_{\fR^k} \|z\|e^{-\|z\|_2^2} \, dz \right) \varepsilon (M+L_f) \\
&\le 2\max\{L_Z, q_Z(0)\} \pi^{-k/2} \left( \int_{\fR^k} \|z\|e^{-\|z\|_2^2} \, dz \right) \varepsilon \sup_{z\in\fR^k} \left( |f(z)| + \|Df(z)\| \right),
\end{align*}
which shows that
\begin{align*}
\left| \mathbb{E}_Z \left[ \left( \tilde{\delta}^\varepsilon(Z) - \delta(Z) \right) f(Z) \right] \right|
& \le \int_{\fR^k} |f(z)q_Z(z) - f(0)q_Z(0)| \tilde{\delta}^\varepsilon(z) \, dz \\
& \le C_\delta \, \varepsilon \sup_{z\in\fR^k} \left( |f(z)| + \|Df(z)\| \right)
\end{align*}
where 
\[
C_\delta = 2\max\{L_Z, q_Z(0)\} \pi^{-k/2} \left( \int_{\fR^k} \|z\|e^{-\|z\|_2^2} \, dz \right).
\]
% \[
% C_1 \stackrel{\Delta}{=} \|q_Z\|_\infty \pi^{-k/2} \int_{\fR^k} \|z\| e^{-\|z\|_2^2} \, dz < \infty.
% \]
% Next, we bound the first integral in \eqref{eqn:fqz_integral_bound}.
% Because $q_Z$ is continuous, there exists $\rho > 0$ such that $|q_Z(z) - q_Z(0)| < \frac{\xi}{3M}$ for all $\|z\| \le \rho$.
% Then, we have
% \begin{align*}
% \int_{\fR^k} M|q_Z(z)-q_Z(0)| \tilde{\delta}^\varepsilon (z) \, dz &= \int_{\|z\|\le \rho} M|q_Z(z)-q_Z(0)| \tilde{\delta}^\varepsilon (z) \, dz + \int_{\|z\| > \rho} M|q_Z(z)-q_Z(0)| \tilde{\delta}^\varepsilon (z) \, dz \\
% &< \frac{\xi}{3} \int_{\|z\|\le\rho} \tilde{\delta}^\varepsilon(z)\,dz + 2M\|q_Z\|_\infty \int_{\|z\|>\rho} \tilde{\delta}^\varepsilon(z) \, dz \\
% &< \frac{\xi}{3} + 2M\|q_Z\|_\infty \int_{\|z\|>\rho} \tilde{\delta}^\varepsilon(z) \, dz.
% \end{align*}
% Because $\int_{\|z\|>\rho} \tilde{\delta}^\varepsilon(z)\,dz \to 0$ as $\varepsilon\downarrow 0$, there exists $\varepsilon'$ such that \begin{align*}
% \int_{\|z\|>\rho} \tilde{\delta}^{\varepsilon'}(z)\,dz < \frac{\xi}{6\|q_Z\|_\infty}.
% \end{align*} 

\end{proof}

%%%%%%%%%%%%%%%%%%%%%%%%%%%% Fourier transform of Gaussian %%%%%%%%%%%%%%%%%%%%%%%%%%%%
%%%%%%%%%%%%%%%%%%%%%%%%%%%% Fourier transform of Gaussian %%%%%%%%%%%%%%%%%%%%%%%%%%%%
\begin{lemma}
\label{lemma:Fourier-Gaussian}
\citep[p.\ 302]{ASbook}
Denote by $\cF[\cdot]$ be the Fourier transform operator.
% Define $\tilde{\delta}^\varepsilon(z) = \frac{\pi^{-k/2}}{\varepsilon^k} e^{-\|z/\varepsilon\|_2^2}$ for $\varepsilon > 0$.
Then
\begin{align*}
\cF[ \Tilde{\delta}^{\varepsilon} ] (\omega) = e^{-\pi^2 \varepsilon^2 \|\omega\|^2}.
\end{align*}
In particular, $\cF[\tilde{\delta}^\varepsilon] (\omega)$ is bounded, and
\begin{gather}
% \label{eqn:Fourier_Gaussian_L1}
\nonumber
\int_{\fR^k} \cF[\Tilde{\delta}^{\varepsilon}] (\omega) \,d\omega < \infty,\\
% \label{eqn:Fourier_Gaussian_gradient_L1}
\nonumber
\int_{\fR^k} \|\omega\| \, \cF[\Tilde{\delta}^{\varepsilon}] (\omega) \,d\omega < \infty.
\end{gather}
\end{lemma}

We first provide a proof when $n=1$, which conveys all important ideas of the proof.
Although the general case involves significantly more complicated notations, it does not essentially differ from the simpler case.

\textbf{Proof for the case $n=1$.}

Let $\varepsilon > 0$ be given.

\textbf{Step 1.} Approximate $\delta (z)$ with $\tilde{\delta}^\varepsilon (z)$ in the sense of Lemma \ref{lemma:w11-delta}.

\textbf{Step 2.} Approximate $\tilde{\delta}^\varepsilon (z)$ with an \emph{infinite} combination of functions in $\cG$.

Because both $\tilde{\delta}^\varepsilon$ and $\cF[\tilde{\delta}^\varepsilon]$ are real-valued and positive, using the inverse Fourier transform, we can write
\begin{align}
    \tilde{\delta}^\varepsilon(z) - \tilde{\delta}^\varepsilon(0) &= \mathrm{Re} \int \left(e^{2\pi i \omega^\intercal z} -1 \right) \cF[\tilde{\delta}^\varepsilon] (\omega) \,d\omega = \int \left( \cos \left(2\pi\omega^\intercal z\right) - 1 \right) \cF[\tilde{\delta}^\varepsilon] (\omega) \,d\omega \label{eqn:Fourier_cosine}
    % &= \int \frac{\cos \left(2\pi\omega^\intercal z + 2\pi \theta(\omega) \right) - \cos 2\pi \theta(\omega)}{\|\omega\|} \|\omega\| \left|\cF[\tilde{\delta}^\varepsilon] (\omega)\right| \,d\omega. \label{eqn:Fourier_cosine_show_convergece}
\end{align}
for any $z\in \fR^k$.
Note that by Lemma~\ref{lemma:Fourier-Gaussian}, the integral \eqref{eqn:Fourier_cosine} is always well-defined.

% We may assume $\varepsilon < 1$ without loss of generality.
Fix a large $R > 0$ satisfying
\[
\int_{\|z\| > R} q_Z(z) \, dz < \frac{\pi^{k/2} \varepsilon^{k+1}}{2}.
\]
% so that
% \[
% \left| \int_{\fR^k} \tilde{\delta}^\varepsilon (z) \, q_Z(z)\, dz - \int_{\|z\| \le R} \tilde{\delta}^\varepsilon (z) \, q_Z(z)\, dz \right|
% = \int_{\|z\| > R} \frac{C}{\varepsilon^k} e^{-\|z/\varepsilon\|_2^2} \, q_Z(z)\, dz
% = \int_{\|z\| > R/\varepsilon} Ce^{-\|z\|_2^2} \, q_Z(z)\, dz < \varepsilon.
% \]
Following \citep[Section~4.2]{Telgarsky2020}, for $\|z\| \le R$, the cosine term in \eqref{eqn:Fourier_cosine} can be rewritten as
\begin{align}
    \nonumber & \cos\left(2\pi \omega^\intercal z \right) - 1 \\
    \nonumber & = \int_0^{\omega^\intercal z} -2\pi \sin (2\pi b) \, db \\
    \label{eqn:cos_to_sin_integral} & = \int_0^{R\|\omega\|} -2\pi \, \mathbf{1}_{\{\omega^\intercal z - b \ge 0\}} (z) \, \sin (2\pi b) \, db  + \int_{-R\|\omega\|}^{0} 2\pi \, \mathbf{1}_{\{\omega^\intercal z - b \le 0\}} (z) \, \sin (2\pi b) \, db.
\end{align}
Let $u_g = \lim_{r\to\infty} \sigma_g (r)$ and $\ell_g = \lim_{r\to -\infty} \sigma_g (r)$.
Then by \hyperlink{cond:AG}{(AG)}, we have
\[
\mathbf{1}_{\{r \ge 0\}}(r) = \lim_{\tau\downarrow 0} \frac{1}{u_g - \ell_g} \left( \sigma_g \left( \frac{r}{\tau} \right) - \ell_g \right)
\]
for $r\ne 0$. Hence we can approximate the step function terms in \eqref{eqn:cos_to_sin_integral} using $\sigma_g$:
\begin{align}
    \label{eqn:step_approx_by_sigmoid}
    & \int_0^{R \|\omega\|} \lim_{\tau \downarrow 0} -\frac{2\pi}{u_g - \ell_g} \left( \sigma_g \left( \frac{\omega^\intercal z - b}{\tau} \right) - \ell_g \right) \sin (2\pi b) \, db + \int_{-R \|\omega\|}^{0} \lim_{\tau \downarrow 0} \frac{2\pi}{u_g-\ell_g} \left( \sigma_g \left( \frac{-\omega^\intercal z + b}{\tau} \right) - \ell_g \right) \sin (2\pi b) \, db.
\end{align}
Plugging \eqref{eqn:step_approx_by_sigmoid} into \eqref{eqn:Fourier_cosine}, we obtain
\begin{align}
& \nonumber \tilde{\delta}^\varepsilon(z) - \tilde{\delta}^\varepsilon(0) \\
& \label{eqn:delta_tilde_integral}
\begin{aligned}
    & = \int \int_0^{R \|\omega\|} \lim_{\tau \downarrow 0} -\frac{2\pi}{u_g-\ell_g} \left( \sigma_g \left( \frac{\omega^\intercal z - b}{\tau} \right) - \ell_g \right) \sin (2\pi b) \, \cF[\tilde{\delta}^\varepsilon] (\omega) \, db \,d\omega \\
    & \quad + \int \int_{-R \|\omega\|}^{0} \lim_{\tau \downarrow 0} \frac{2\pi}{u_g - \ell_g} \left( \sigma_g \left( \frac{-\omega^\intercal z + b}{\tau} \right) - \ell_g \right) \sin (2\pi b) \, \cF[\tilde{\delta}^\varepsilon] (\omega) \, db \,d\omega
\end{aligned}
\end{align}
for $\|z\| \le R$.

Observe that because $\sigma_g$ is bounded and by Lemma~\ref{lemma:Fourier-Gaussian}, for any $\tau > 0$ and $z \in \fR^k$,
\begin{align*}
& \int \int_0^{R\|\omega\|} \left| \frac{2\pi}{u_g-\ell_g} \left( \sigma_g \left( \frac{\omega^\intercal z - b}{\tau} \right) - \ell_g \right) \sin (2\pi b) \, \cF[\tilde{\delta}^\varepsilon] (\omega) \right| \, db \,d\omega \le \int  \frac{2\pi \left(\|\sigma_g\|_\infty + \ell_g \right)}{u_g - \ell_g} R \|\omega\| \, \cF[\tilde{\delta}^\varepsilon] \, d\omega < \infty.
\end{align*}
Therefore, by Lebesgue's dominated convergence theorem, we can freely change the order of integration and limit in \eqref{eqn:delta_tilde_integral}.
Using this fact, and applying change of variables, we can rewrite $\tilde{\delta}^\varepsilon(z)$ as
\begin{align*}
\tilde{\delta}^\varepsilon(z) & = \tilde{\delta}^\varepsilon(0) + \lim_{\tau \downarrow 0} \int \int_0^{R \|\omega\|}  -\frac{2\pi}{u_g-\ell_g} \left( \sigma_g \left( \frac{\omega^\intercal z - b}{\tau} \right) - \ell_g \right) \sin (2\pi b) \, \cF[\tilde{\delta}^\varepsilon] (\omega) \, db \,d\omega \\
& \quad \quad \quad \,\,\, + \lim_{\tau \downarrow 0} \int \int_{-R \|\omega\|}^{0} \frac{2\pi}{u_g - \ell_g} \left( \sigma_g \left( \frac{-\omega^\intercal z + b}{\tau} \right) - \ell_g \right) \sin (2\pi b) \, \cF[\tilde{\delta}^\varepsilon] (\omega) \, db \,d\omega \\
& = \theta^\varepsilon_1 \phi (z;\kappa_1) + \lim_{\tau \downarrow 0} \int \int_{-R \|\omega\|}^{0}  -\frac{2\pi \tau^{k+1}}{u_g-\ell_g} \sigma_g \left( \omega^\intercal z + b  \right) \sin (-2\pi \tau b) \, \cF[\tilde{\delta}^\varepsilon] (\tau\omega) \, db \,d\omega \\
& \quad \quad \quad \,\,\, + \lim_{\tau \downarrow 0} \int \int_{-R \|\omega\|}^{0} \frac{2\pi \tau^{k+1}}{u_g - \ell_g} \sigma_g \left( \omega^\intercal z + b \right) \sin (2\pi \tau b) \, \cF[\tilde{\delta}^\varepsilon] (- \tau \omega) \, db \,d\omega \\
& = \theta^\varepsilon_1 \phi (z;\kappa_1) + \lim_{\tau\downarrow 0} \int_{\fR^k \times \fR} \phi (z;\kappa) m_\tau(\kappa) \, d\kappa
\end{align*}
for $\|z\| \le R$.
We specify the notations that were newly introduced.
First, we denoted $\kappa = (\omega, b)$, so that $\phi(z;\kappa) = \sigma_g (\omega^\intercal z + b)$ (note that because we have assumed $n=1$, the generator parameter has dimension $k+1$), and $d\kappa$ is the Lebesgue measure on $\fR^k \times \fR$.
Next, we set $\kappa_1 = (0,b_1)$ with some fixed $b_1 \in \fR$ satisfying $\phi(z;\kappa_1) \equiv \sigma_g (b_1) \ne 0$ and
\begin{align*}
\theta^\varepsilon_1 & = \frac{1}{\sigma_g(b_1)} \left( \tilde{\delta}^\varepsilon(0) + \int \int_0^{R\|\omega\|} \frac{2\pi \ell_g}{u_g - \ell_g} \sin(2\pi b) \cF[\tilde{\delta}^\varepsilon](\omega) \, db\,d\omega - \int \int_{-R\|\omega\|}^{0} \frac{2\pi \ell_g}{u_g - \ell_g} \sin(2\pi b) \cF[\tilde{\delta}^\varepsilon](\omega) \, db\,d\omega \right) \in \fR. % \\ & = \frac{1}{\sigma_g (b_1)} \tilde{\delta}^\varepsilon (0) \in \fR.
\end{align*}
Finally, we define the density function $m_\tau(\kappa)$ as
\begin{align}
\nonumber
m_\tau(\kappa) &= \frac{2\pi\tau^{k+1}}{u_g - \ell_g} \left( - \sin (-2\pi \tau b) \, \cF[\tilde{\delta}^\varepsilon](\tau\omega) \, \mathbf{1}_{\{-R\|\omega\| \le b \le 0\}} (\kappa)
+ \sin (2\pi \tau b) \, \cF[\tilde{\delta}^\varepsilon] (-\tau\omega) \, \mathbf{1}_{\{-R\|\omega\| \le b \le 0\}} (\kappa) \right) \\
\label{eqn:m_eta}
&= \frac{4\pi \tau^{k+1}}{u_g-\ell_g} e^{-\pi^2 \varepsilon^2 \tau^2 \|\omega\|^2} \sin(2\pi\tau b) \, \mathbf{1}_{\{-R\|\omega\| \le b \le 0\}} (\kappa),
\end{align}
where we used Lemma \ref{lemma:Fourier-Gaussian} to obtain the second equality.

Now we bound the error in using the expression \eqref{eqn:cos_to_sin_integral} in the case $\|z\| > R$.
Observe that
\begin{align*}
& \cos (2\pi\omega^\intercal z) - 1 - \int_0^{R\|\omega\|} -2\pi \, \mathbf{1}_{\{\omega^\intercal z - b \ge 0\}} (z) \, \sin (2\pi b) \, db  + \int_{-R\|\omega\|}^{0} 2\pi \, \mathbf{1}_{\{\omega^\intercal z - b \le 0\}} (z) \, \sin (2\pi b) \, db \\
& \quad = \left(\cos (2\pi\omega^\intercal z) - \cos (2\pi R\|\omega\|) \right) \mathbf{1}_{\{|\omega^\intercal z| > R\|\omega\|\}}(\omega),
\end{align*}
and thus
\begin{align}
\label{eqn:pointwise_convergence_final_form}
\tilde{\delta}^\varepsilon(z) - \theta_1^\varepsilon \phi(z;\kappa_1) - \lim_{\tau\downarrow 0} \int_{\fR^k \times \fR} \phi(z;\kappa) m_\tau(\kappa)\, d\kappa
= \int_{\{\omega\,|\, |\omega^\intercal z| > R\|\omega\|\}} \left(\cos (2\pi\omega^\intercal z) - \cos (2\pi R\|\omega\|) \right) \cF[\tilde{\delta}^\varepsilon](\omega)  \, d\omega
\end{align}
for all $z \in \fR^k$.
The defining equation \eqref{eqn:m_eta} shows that $m_\tau$ is bounded and $m_\tau \in L^1 (d\kappa)$ with
\begin{align*}
\int_{\fR^k \times \fR} \left|m_\tau(\kappa)\right|\, d\kappa \le \frac{4\pi R}{u_g-\ell_g} \int_{\fR^k} \tau^{k} \|\tau \omega\| e^{-\pi^2 \varepsilon^2 \|\tau\omega\|^2} \, d\omega
= \frac{4\pi R}{u_g - \ell_g} \int_{\fR^k} \|\omega\| e^{-\pi^2\varepsilon^2 \|\omega\|^2} \, d\omega.
\end{align*}
Therefore, the family
\begin{align*}
\left\{ \tilde{\delta}^\varepsilon(z) - \theta^\varepsilon_1 \phi(z;\kappa_1) - \int_{\fR^k \times \fR} \phi (z;\kappa) f(z) m_\tau (\kappa) \, d\kappa\right\}_{\tau > 0}
\end{align*}
is uniformly bounded.
Applying the dominated convergence theorem to the pointwise convergence result \eqref{eqn:pointwise_convergence_final_form} with respect to the probability measure $q_{Z}(z)\,dz$, we obtain
\begin{align*}
& \lim_{\tau\downarrow 0} \mathbb{E}_{Z} \left[ \left| \tilde{\delta}^\varepsilon(Z) - \theta^\varepsilon_1 \phi(Z;\kappa_1) - \int_{\fR^k \times \fR} \phi (Z;\kappa) m_\tau (\kappa) \, d\kappa \right| \right] \\
& = \mathbb{E}_Z \left[\left| \int_{\{\omega\,|\, |\omega^\intercal Z| > R\|\omega\|\}} \left(\cos (2\pi\omega^\intercal Z) - \cos (2\pi R\|\omega\|) \right) \cF[\tilde{\delta}^\varepsilon](\omega)  \, d\omega \right|\right] \\
& \le \mathbb{E}_Z \left[ \int_{\{\omega\,|\, |\omega^\intercal Z| > R\|\omega\|\}} 2 \cF[\tilde{\delta}^\varepsilon](\omega)  \, d\omega \right] \\
& \le \mathbb{E}_Z \left[ \mathbf{1}_{\{\|z\| > R\}} (Z) \int_{\fR^k} 2\cF[\tilde{\delta}^\varepsilon](\omega) \, d\omega \right] \\
& = \left( \int_{\{\|z\| > R\}} q_Z(z) \, dz \right) \left( \int_{\fR^k} 2\cF[\tilde{\delta}^\varepsilon](\omega) \, d\omega \right) < \frac{\pi^{k/2} \varepsilon^{k+1}}{2} \frac{2}{\pi^{k/2} \varepsilon^k} = \varepsilon.
\end{align*}

\textbf{Step 3.} 
Approximate the integral over $\fR^{k}\times \fR$ by an integral over a ball of finite radius.

We fix some $\tau = \tau(\varepsilon)$ satisfying $\mathbb{E}_{Z} \left[ \left| \tilde{\delta}^\varepsilon(Z) - \theta^\varepsilon_1 \phi(Z;\kappa_1) - \int_{\fR^k \times \fR} \phi (Z;\kappa) m_\tau (\kappa) \, d\kappa \right| \right] < 2 \varepsilon$.
Because $\sigma_g$ is bounded and $m_\tau \in L^1 (d\kappa)$, there exists $K>0$ large enough so that
\begin{align*}
    \int_{\|\kappa\| > K} |m_\tau (\kappa)| \, d\kappa < \frac{\varepsilon}{\|\sigma_g\|_\infty}.
\end{align*}
Then for any bounded continuous function $f:\fR^k \to \fR$ we have
\begin{align*}
& \mathbb{E}_{Z} \left[ \left| \tilde{\delta}^\varepsilon(Z)f(Z) - \theta^\varepsilon_1 \phi(Z;\kappa_1) f(Z) - \int_{\|\kappa\| \le K} \phi (Z;\kappa) f(Z) m_\tau (\kappa) \, d\kappa \right| \right]\\
& \le \mathbb{E}_{Z} \left[ \left| \tilde{\delta}^\varepsilon(Z)f(Z) - \theta^\varepsilon_1 \phi(Z;\kappa_1) f(Z) - \int_{\fR^k \times \fR} \phi (Z;\kappa) f(Z) m_\tau (\kappa) \, d\kappa \right| \right] \\
& \quad \quad + \mathbb{E}_Z \left[ \int_{\|\kappa\| > K} \left| \phi (Z;\kappa) f(Z) m_\tau (\kappa) \right|\, d\kappa \right] \\
& \le 2 \varepsilon \|f\|_\infty + \|\sigma_g\|_\infty \|f\|_\infty \mathbb{E}_Z \left[ \int_{\|\kappa\| > K} |m_\tau(\kappa)|\, d\kappa \right] \le 3\varepsilon \|f\|_\infty.
\end{align*}

\textbf{Step 4.}
Approximate the integral over a finite ball by a finite linear combination of random functions in $\cG$.

Define
\[
m_{\tau,K}(\kappa) = \begin{cases}
    m_\tau (\kappa) & \text{if } \|\kappa\| \le K, \\
    0               & \text{otherwise}.
\end{cases}
\]
Denote by $p(\kappa)$ the strictly positive continuous density function from which we randomly sample the generator parameters.

Note that we have
\begin{align*}
    C_K \stackrel{\Delta}{=} \sup_{\kappa} \left| \frac{m_{\tau,K} (\kappa)}{p(\kappa)} \right| < \infty
\end{align*}
because $\|m_\tau\|_\infty < \infty$ and $1/p(\kappa)$ is bounded over a compact set.

Now, rewrite the integral from Step 3 as
\begin{align*}
\int_{\|\kappa\|\le K} \phi \left( z;\kappa \right) \, m_{\tau} (\kappa)\,d\kappa = \int \phi \left( z;\kappa \right) \frac{m_{\tau,K} (\kappa)}{p(\kappa)} \, p(\kappa)\,d\kappa.
\end{align*}
We will show that if we sample $\kappa_2, \dots, \kappa_{N_g}$ (IID) according to $p(\kappa)$, then for sufficiently large $N_g$,
\begin{align*}
\int \phi \left( Z;\kappa \right) \frac{m_{\tau,K} (\kappa)}{p(\kappa)} \, p(\kappa)\,d\kappa \approx \frac{1}{N_g-1} \sum_{i=2}^{N_g} \phi (Z;\kappa_i) \frac{m_{\tau,K}(\kappa_i)}{p(\kappa_i)}
\end{align*}
with high probability over $\kappa_2, \dots, \kappa_{N_g}$.
(The indexing begins with $i=2$ because $\kappa_1$ is reserved for the constant function.)
When we draw each $\kappa_i$, we are in fact sampling the corresponding function
\begin{align*}
h_i \stackrel{\Delta}{=} \frac{m_{\tau,K}(\kappa_i)}{p(\kappa_i)} \phi (\cdot;\kappa_i) \in \cH \stackrel{\Delta}{=} L^2 (q_Z(z)\,dz).
\end{align*}
Indeed, $h_i$ are uniformly bounded with $\|h_i\|_\infty \le \|\sigma_g\|_\infty C_K$ for all $i=2,\dots,N_g$, which implies $\|h_i\|_\cH \le \|\sigma_g\|_\infty C_K$.
That is, $\frac{m_{\tau,K}(\kappa)}{p(\kappa)} \phi (\cdot;\kappa)$ is a bounded random variable with random realizations in $\cH$.
Also, we have
\begin{align*}
\mathbb{E}_{\kappa \sim p(\kappa)} \left[ \frac{m_{\tau,K}(\kappa)}{p(\kappa)} \phi (\cdot;\kappa) \right] = \int \phi (\cdot;\kappa) \frac{m_{\tau,K}(\kappa)}{p(\kappa)} p(\kappa) \, d\kappa = \int_{\|\kappa\| \le K} \phi (\cdot;\kappa) \, m_\tau (\kappa)\, d\kappa.
\end{align*}
Therefore, applying the McDiarmid-type bound from \citep[Lemma~4]{rahimi2007random}, we get
\begin{align}
\label{eqn:integral_to_finite_sum_Hilbert}
\begin{aligned}
    \left\| \frac{1}{N_g-1}\sum_{i=2}^{N_g} h_i - \mathbb{E}_{\kappa \sim p(\kappa)} \left[ \frac{m_{\tau,K}(\kappa)}{p(\kappa)} \phi (\cdot;\kappa) \right] \right\|_\cH &= 
    \left\| \sum_{i=2}^{N_g} \frac{m_{\tau,K}(\kappa_i)}{(N_g-1)p(\kappa_i)} \phi (\cdot;\kappa_i) - \int_{\|\kappa\| \le K} \phi (\cdot;\kappa) \, m_\tau (\kappa)\, d\kappa \right\|_\cH \\
    & \le \frac{\|\sigma_g\|_\infty C_K}{\sqrt{N_g-1}} \left( 1 + \sqrt{2 \log \frac{1}{\zeta}} \right),
\end{aligned}
\end{align}
with probability at least $1-\zeta$ over $\kappa_2, \dots, \kappa_{N_g}$.
Fix $N_g$ large enough so that the right hand side of \eqref{eqn:integral_to_finite_sum_Hilbert} is less than $\varepsilon$, and let $\theta^\varepsilon_i = \frac{m_{\tau,K}(\kappa_i)}{(N_g-1)p(\kappa_i)}$ for $i=2,\dots,N_g$.
Then, using Jensen's inequality, we obtain
\begin{align*}
\varepsilon > \left\| \sum_{i=2}^{N_g} \theta^\varepsilon_i \phi(\cdot;\kappa_i) - \int_{\|\kappa\| \le K} \phi(\cdot;\kappa) \, m_\tau(\kappa)\, d\kappa \right\|_\cH &= \left( \mathbb{E}_Z \left[ \left| \sum_{i=2}^{N_g} \theta^\varepsilon_i \phi(Z;\kappa_i) - \int_{\|\kappa\| \le K} \phi(Z;\kappa) \, m_\tau(\kappa)\, d\kappa \right|^2 \right] \right)^{1/2} \\
& \ge \mathbb{E}_Z \left[ \left| \sum_{i=2}^{N_g} \theta^\varepsilon_i \phi(Z;\kappa_i) - \int_{\|\kappa\| \le K} \phi(Z;\kappa) \, m_\tau(\kappa)\, d\kappa \right| \right]
\end{align*}
with probability $\ge 1-\zeta$.

\textbf{Step 5.} Combine Steps 1 through 4.

Let $\kappa_1,\dots,\kappa_{N_g}$ be as above, and $\phi_i(z) = \phi(z;\kappa_i)$.
For any continuously differentiable function $f:\fR^k \to \fR$ such that $\sup_{z\in\fR^k} \left(|f(z)| + \|\nabla f(z)\|\right) < \infty$,
with probability at least $1-\zeta$, we have
\begin{align*}
    & \left| \mathbb E_Z \left[ \left( \delta (Z) - \sum_{i=1}^{N_g} \theta_i^{\varepsilon} \phi_i (Z) \right) f(Z)  \right] \right| \\
    & \le \EZ{ \left|\left(\delta(Z) - \tilde{\delta}^{\varepsilon}(Z)\right) f(Z) \right|} + \mathbb{E}_Z \left[ \left| \tilde{\delta}^\varepsilon(Z)f(Z) - \theta^\varepsilon_1 \phi_1(Z) f(Z) - \int_{\|\kappa\| \le K} \phi (Z;\kappa) f(Z) \, m_\tau (\kappa) \, d\kappa \right| \right] \\
    & \quad + \mathbb{E}_Z \left[ \left|  \theta^\varepsilon_1 \phi_1(Z) f(Z) + \int_{\|\kappa\| \le K} \phi(Z;\kappa) f(Z) \, m_\tau(\kappa)\, d\kappa - \sum_{i=1}^{N_g} \theta^\varepsilon_i \phi_i (Z) f(Z) \right| \right] \\
    & \le C_\delta \, \varepsilon \sup_{z\in\fR^k} \left( |f(z)| + \|\nabla f(z)\| \right) + 3 \varepsilon \|f\|_\infty + \|f\|_\infty \mathbb{E}_Z \left[ \left| \int_{\|\kappa\| \le K} \phi(Z;\kappa) \, m_\tau(\kappa)\, d\kappa - \sum_{i=2}^{N_g} \theta^\varepsilon_i \phi_i(Z) \right| \right] \\
    & \le (C_\delta + 4) \, \varepsilon \sup_{z\in\fR^k} \left( \|f(z)\| + \|\nabla f(z)\| \right),
\end{align*}
where $C_\delta$ is the constant (not depending on $\varepsilon$) defined in Lemma~\ref{lemma:w11-delta}.
This completes the proof for the case $n=1$.

\textbf{Proof for the general case $n>1$.}

The crux of the general case is that $\kappa$ cannot be sampled coordinate-wisely, but we must keep only one coordinate active, while suppressing the others.
To achieve this, we simply accept $\kappa$'s whose rows are negligibly small except possibly for the $l$-th row.
We express $\kappa \in \fR^{n\times k} \times \fR^n$ in the form
\[
\kappa =
\begin{bmatrix}
\kappa^{(1)} \\
\vdots \\
\kappa^{(n)}
\end{bmatrix} = \begin{bmatrix}
\left(\omega^{(1)}\right)^\intercal & b^{(1)} \\
\vdots & \vdots \\
\left(\omega^{(n)}\right)^\intercal & b^{(n)} 
\end{bmatrix},
\]
where $\omega^{(j)} \in \fR^k$, $b^{(j)} \in \fR$ and $\kappa^{(j)} = \left[\left(\omega^{(j)}\right)^\intercal, b^{(j)}\right]$ for $j=1,\dots,n$.

Fix $1 \le l \le n$ and $\varepsilon > 0$.
Define
\[
\tilde{\delta}^{(\varepsilon,l)} (z) \stackrel{\Delta}{=} \begin{bmatrix}
0 \\
\vdots \\
0 \\
\tilde{\delta}^\varepsilon (z) \\
0 \\
\vdots \\
0
\end{bmatrix}
\leftarrow\text{on $l$\nobreakdash-th coordinate.}
\]
Let
\begin{gather*}
C_m \stackrel{\Delta}{=} \frac{4}{u_g - \ell_g} \int_{\fR^k} e^{-\pi^2\varepsilon^2 \|\omega\|^2} \, d\omega,
% \\ C_{|m|} \stackrel{\Delta}{=} \frac{4\pi}{u_g - \ell_g} \int_{\fR^k} \|\omega\| e^{-\pi^2\varepsilon^2 \|\omega\|^2} \, d\omega,
\end{gather*}
which is a constant depending only on $\varepsilon$.
Take a large $R>0$ satisfying
\[
\int_{\|z\|>R} q_Z(z) \, dz < \min \left\{ \frac{\pi^{k/2}\varepsilon^{k+1}}{2} , \frac{\varepsilon}{4(n-1)\|\sigma_g\|_\infty C_m} \right\}.
\]
From Steps 2 and 3 in the case $n=1$, we can find a density function $m = m_\tau$ of the form
\[
m_\tau (\kappa^{(l)}) = \frac{4\pi \tau^{k+1}}{u_g-\ell_g} e^{-\pi^2 \varepsilon^2 \tau^2 \|\omega^{(l)}\|^2} \sin(2\pi\tau b^{(l)}) \, \mathbf{1}_{\{-R\|\omega^{(l)}\| \le b^{(l)} \le 0\}} (\kappa^{(l)})
\]
on $\fR^k \times \fR$ (with $\tau = \tau(\varepsilon) < 1$)
such that for some $\rho^{(l)} \in \fR$ and $K>0$ large enough,
\begin{gather*}
\mathbb{E}_{Z} \left[ \left| \tilde{\delta}^\varepsilon(Z) - \rho^{(l)} - \int_{\|\kappa^{(l)}\| \le K} \sigma_g \left( (\omega^{(l)})^\intercal Z + b^{(l)} \right) m (\kappa^{(l)}) \, d\kappa^{(l)} \right| \right] < 3 \varepsilon. 
\end{gather*}
% Recall that
% \[
% \int_{\fR^{k}\times \fR} |m(\kappa^{(l)})|\,d\kappa^{(l)} \le C_{|m|} R.
% \]
Note that we can bound
\begin{align*}
\left| \int_{\|\kappa^{(l)}\| \le K} m(\kappa^{(l)}) \, d\kappa^{(l)} \right|
&= \left| \int_{\|\omega^{(l)}\|\le K} \frac{2\tau^{k+1}}{u_g-\ell_g} e^{-\pi^2 \varepsilon^2 \tau^2 \|\omega^{(l)}\|^2} \int_{-\min\{R\|\omega^{(l)}\|, K-\|\omega^{(l)}\|\}}^0 2\pi \sin(2\pi b^{(l)}) \, db^{(l)} \, d\omega^{(l)} \right |
\\
&= \left| \int_{\|\omega^{(l)}\|\le K} \frac{2\tau^{k+1}}{u_g-\ell_g} e^{-\pi^2 \varepsilon^2 \tau^2 \|\omega^{(l)}\|^2} \left( \cos\left(2\pi\min\{R\|\omega^{(l)}\|, K-\|\omega^{(l)}\|\}\right) - 1 \right) \, d\omega^{(l)}
\right|\\
& \le \tau \int_{\fR^k} \frac{4}{u_g - \ell_g} e^{-\pi^2 \varepsilon^2 \tau^2 \|\omega^{(l)}\|^2} \tau^k \, d\omega^{(l)} = \tau C_m < C_m.
\end{align*}
For $\xi > 0$, consider the set 
\[
\cK_\xi^{(l)} \stackrel{\Delta}{=}
\left\{
\kappa \in \fR^{n\times k}\times \fR^n \, \big| \, \|\kappa^{(l)}\| \le K, \, \|\kappa^{(j)}\| \le \xi \text{ for } j \ne l
\right\}.
\]
Denote by $\cB_\xi$ the closed ball of radius $\xi$ in $\fR^{k+1}$, centered at $0$.
Now define
\[
m_\xi^{(l)} \left(\kappa^{(1)}, \cdots, \kappa^{(n)}\right) \stackrel{\Delta}{=}
m(\kappa^{(l)}) \frac{ \mathbf{1}_{\cK_\xi^{(l)}} \left(\kappa^{(1)}, \cdots, \kappa^{(n)}\right) } { \Vol \left(\cB_\xi\right)^{n-1} }.
\]
We will show that for sufficiently small $\xi$ and some constant vector $\bv^{(\varepsilon,l)} \in \fR^n$,
\[
\mathbb{E}_{Z} \left[ \left\| \tilde{\delta}^{(\varepsilon,l)}(Z) - \bv^{(\varepsilon,l)} - \int_{\fR^{n\times k}\times \fR^n} \phi (Z;\kappa) m_\xi^{(l)} (\kappa) \, d\kappa \right\|_2 \right] = \mathcal{O}(\varepsilon).
\]
Note that given $z \in \fR^k$,
\[
\tilde{\Phi}_\xi^{(l)} (z) \stackrel{\Delta}{=} \int_{\fR^{n\times k}\times \fR^n} \phi (z;\kappa) m_\xi^{(l)} (\kappa) \, d\kappa = 
\begin{bmatrix}
\int_{\fR^{n\times k} \times \fR^n} \sigma_g \left( (\omega^{(1)})^\intercal z + b^{(l)} \right) m_\xi^{(l)}(\kappa) \, d\kappa \\
\vdots \\
\int_{\fR^{n\times k} \times \fR^n} \sigma_g \left( (\omega^{(n)})^\intercal z + b^{(l)} \right) m_\xi^{(l)}(\kappa) \, d\kappa
\end{bmatrix} \in \fR^n.
\]
For $j=1,\dots,n$, we denote the $j$-th component function of $\tilde{\Phi}_\xi^{(l)}$ by $\left[\tilde{\Phi}_\xi^{(l)}\right]_j$.
Observe that if we denote $d\kappa^{(-l)} = d\kappa^{(1)}\cdots d\kappa^{(l-1)} d\kappa^{(l+1)} \cdots d\kappa^{(n)}$, then by our construction of $m$ and $K$,
\begin{align*}
\left[\tilde{\Phi}_\xi^{(l)}\right]_l (Z) &= \int_{\fR^{k}\times \fR} \sigma_g \left( (\omega^{(l)})^\intercal Z + b^{(l)} \right) m(\kappa^{(l)}) \left( \int_{\fR^{(n-1)\times k} \times \fR^{n-1}} \frac{ \mathbf{1}_{\cK_\xi^{(l)}} \left(\kappa^{(1)}, \cdots, \kappa^{(n)}\right) } { \Vol \left(\cB_\xi\right)^{n-1} } \, d\kappa^{(-l)} \right) d\kappa^{(l)} \\
& = \int_{\|\kappa^{(l)}\| \le K} \sigma_g \left( (\omega^{(l)})^\intercal Z + b^{(l)} \right) m (\kappa^{(l)}) \, d\kappa^{(l)},
\end{align*}
which is $3\varepsilon$-close to $\tilde{\delta}^\varepsilon (Z) - \rho^{(l)}$ within $L^1(q_Z(z)\,dz)$, regardless of $\xi$.

Next we bound the remaining components of $\tilde{\Phi}_\xi^{(l)}$.
Since $\sigma_g (r)$ is continuous at $r=0$ (by \hyperlink{cond:AG}{(AG)}), we can take $\xi$ so that
\begin{equation}
\label{eqn:sigma_g_continuity}
|\sigma_g(r) - \sigma_g(0)| < \frac{\varepsilon}{2(n-1) C_{m}}
\end{equation}
holds for all $|r| < (1+R)\xi$.
Observe that for $j\ne l$,
\begin{align*}
\left[\tilde{\Phi}_\xi^{(l)}\right]_j (z) &= \int_{\fR^{(n-1)\times k}\times \fR^{n-1}} \sigma_g \left( (\omega^{(j)})^\intercal z + b^{(j)} \right) \frac{ \prod_{m\ne l} \mathbf{1}_{\cB_\xi} (\kappa^{(m)}) } { \Vol \left(\cB_\xi\right)^{n-1} }
\left( \int_{\|\kappa^{(l)}\|\le K} m(\kappa^{(l)}) \, d\kappa^{(l)} \right) d\kappa^{(-l)} \\
&=  \frac{1} { \Vol \left(\cB_\xi\right)} \int_{\|\kappa^{(j)}\| \le \xi} 
\sigma_g \left( (\omega^{(j)})^\intercal z + b^{(j)} \right) 
\left( \int_{\|\kappa^{(l)}\|\le K} m(\kappa^{(l)}) \, d\kappa^{(l)} \right) d\kappa^{(j)}.
\end{align*}
Define
\[
\rho^{(-l)} \stackrel{\Delta}{=} \int_{\|\kappa^{(l)}\|\le K} \sigma_g (0) m(\kappa^{(l)}) \, d\kappa^{(l)} = \frac{1} { \Vol \left(\cB_\xi\right)} \int_{\|\kappa^{(j)}\| \le \xi} 
\sigma_g (0) \left( \int_{\|\kappa^{(l)}\|\le K} m(\kappa^{(l)}) \, d\kappa^{(l)} \right) d\kappa^{(j)}.
\]
Then we have
\begin{align*}
\left| \left[\tilde{\Phi}_\xi^{(l)}\right]_j (z) - \rho^{(-l)} \right| & \le 
\frac{1} { \Vol \left(\cB_\xi\right)} \int_{\|\kappa^{(j)}\| \le \xi} 
\left| \sigma_g \left( (\omega^{(j)})^\intercal z + b^{(j)} \right) - \sigma_g (0) \right|
\left| \int_{\|\kappa^{(l)}\|\le K} m(\kappa^{(l)}) \, d\kappa^{(l)} \right| d\kappa^{(j)} \\
& \le \frac{1} { \Vol \left(\cB_\xi\right)} \int_{\|\kappa^{(j)}\| \le \xi} 
C_m \left| \sigma_g \left( (\omega^{(j)})^\intercal z + b^{(j)} \right) - \sigma_g (0) \right| d\kappa^{(j)}.
\end{align*}
Note the integrand is nonzero only when $\|\kappa^{(j)}\| \le \xi$, which implies $\|\omega^{(j)}\|, |b^{(j)}| \le \xi$.
Therefore, on the event $\|z\|\le R$, we have $\left|(\omega^{(j)})^\intercal z + b^{(j)}\right| \le \xi (1+\|z\|) \le \xi(1+R)$, so \eqref{eqn:sigma_g_continuity} gives
\[
\left| \left[\tilde{\Phi}_\xi^{(l)}\right]_j (z) - \rho^{(-l)} \right| \le C_m \frac{\varepsilon}{2(n-1)C_m} = \frac{\varepsilon}{2(n-1)}.
\]
When $\|z\| > R$, the crude bound
\[
\left| \left[\tilde{\Phi}_\xi^{(l)}\right]_j (z) - \rho^{(-l)} \right| \le 2\|\sigma_g\|_\infty C_m
\]
is enough, because $\mathrm{Prob}_Z[\|Z\| \ge R] < \frac{\varepsilon}{4(n-1)\|\sigma_g\|_\infty C_m}$.
We have established
\begin{align*}
\mathbb{E}_Z \left[ \left|
\left[\tilde{\Phi}_\xi^{(l)}\right]_j (Z) - \rho^{(-l)} \right| \right] < \frac{\varepsilon}{n-1} 
\end{align*}
for all $j\ne l$.

Now, with
\[
\bv^{(\varepsilon,l)} = \begin{bmatrix}
-\rho^{(-l)} \\
\vdots \\
-\rho^{(-l)} \\
\rho^{(l)} \\
-\rho^{(-l)} \\
\vdots \\
-\rho^{(-l)}
\end{bmatrix}
\leftarrow\text{on $l$\nobreakdash-th coordinate,}
\]
we have
\begin{align*}
& \mathbb{E}_{Z} \left[ \left\| \tilde{\delta}^{(\varepsilon,l)}(Z) - \bv^{(\varepsilon,l)} - \int_{\fR^{n\times k}\times \fR^n} \phi (Z;\kappa) m_\xi^{(l)} (\kappa) \, d\kappa \right\|_2 \right]\\
& \le \mathbb{E}_{Z} \left[ \left\| \tilde{\delta}^{(\varepsilon,l)}(Z) - \bv^{(\varepsilon,l)} - \int_{\fR^{n\times k}\times \fR^n} \phi (Z;\kappa) m_\xi^{(l)} (\kappa) \, d\kappa \right\|_1 \right] \\
& = \mathbb{E}_Z \left[\left| \tilde{\delta}^\varepsilon (Z) - \rho^{(l)} - \left[\tilde{\Phi}_\xi^{(l)}\right]_l (z) \right|\right] + \sum_{j\ne l} \mathbb{E}_Z \left[ \left|
\left[\tilde{\Phi}_\xi^{(l)}\right]_j (Z) - \rho^{(-l)} \right| \right] \\
& < 3\varepsilon + (n-1) \frac{\varepsilon}{n-1}  = 4\varepsilon.
\end{align*}

The space of vector functions $h = ([h]_1,\dots,[h]_n) \colon \fR^k \to \fR^n$ satisfying $\EZ{|[h]_j(Z)|^2} < \infty$ for each $j=1,\dots,n$ can be identified as the direct sum of $L^2$ spaces
\[
\cH \stackrel{\Delta}{=} \bigoplus_{j=1}^n L^2(q_Z(z) \, dz).
\]
This is a Hilbert space equipped with the inner product $\langle g,h \rangle_\cH = \sum_{j=1}^n \mathbb{E}_Z [[g]_j(Z)[h]_j(Z)] = \mathbb{E}_Z [g^\intercal (Z) h(Z)]$.
Now let $p(\kappa) > 0$ be the density function on $\fR^{n\times k} \times \fR^n$ from which we sample $\kappa$'s, and define
\[
C_\cK^{(l)} \stackrel{\Delta}{=} \sup_{\kappa} \frac{m_\xi^{(l)}(\kappa)}{p(\kappa)},
\]
which is finite because $m_\xi^{(l)}$ is bounded and compactly supported, while $p$ is positive and continuous.
% where $p(\kappa) > 0$ denotes the density function on $\fR^{n\times k} \times \fR^n$ that we are sampling from.
For each random $\kappa_i$, $i=n+1,\dots,N_g$, the corresponding realization 
\[
h_i := \frac{m_\xi^{(l)}(\kappa_i)}{p(\kappa_i)} \, \phi(\cdot; \kappa_i) \in \cH
\]
satisfies $\|h_i\|_\cH \le \sqrt{n} \|\sigma_g\|_\infty C_\cK^{(l)}$.
Hence, as in the $n=1$ case,
\begin{align*}
    \left\| \frac{1}{N_g-n}\sum_{i=n+1}^{N_g} h_i - \int_{\fR^{n\times k}\times \fR^n} \phi (Z;\kappa) m_\xi^{(l)} (\kappa) \, d\kappa \right\|_\cH \le \frac{\sqrt{n} \|\sigma_g\|_\infty C_\cK^{(l)}}{\sqrt{N_g-n}} \left( 1 + \sqrt{2 \log \frac{1}{\zeta}} \right)
\end{align*}
with probability $\ge 1-\zeta$ over $\kappa_{n+1}, \dots, \kappa_{N_g}$.
Let $\theta^{(\varepsilon,l)}_i = \frac{m_\xi^{(l)}(\kappa_i)}{(N_g-n) p(\kappa_i)}$ for $i=n+1,\dots,N_g$.
Take $\kappa_1 = (0_{n\times k},b_1), \dots, \kappa_n = (0_{n\times k},b_n)$, where $b_1,\dots,b_n \in \fR^n$, in a way that the constant vectors $\sigma_g(b_i)$ are linearly independent.
Then there exist $\theta^{(\varepsilon,l)}_1, \dots, \theta^{(\varepsilon,l)}_n \in \fR$ such that
\begin{align*}
& \sum_{i=1}^n \theta^{(\varepsilon,l)}_i \sigma_g (b_i) = \sum_{i=1}^n \theta^{(\varepsilon,l)}_i \phi(z;\kappa_i) = \bv^{(\varepsilon,l)}.
\end{align*}
Given $f \colon \fR^k \to \fR^n$, let $M = \sup_{z\in\fR^k} \|f(z)\|_2$.
%and $L = \sup_{z\in\fR^k} \|Df(z)\|$.
Chaining all the approximation steps, we have
\begin{align*}
& \left|\mathbb{E}_{Z} \left[ \left( \delta^{(l)}(Z) - \sum_{i=1}^{N_g} \theta^{(\varepsilon,l)}_i \phi(Z;\kappa_i) \right)^\intercal f(Z) \right] \right| \\
& \le \left| \mathbb{E}_Z \left[ \left( \delta^{(l)}(Z) - \tilde{\delta}^{(\varepsilon,l)} (Z)\right)^\intercal f(Z)  \right] \right| + \mathbb{E}_Z \left[ \left| \left( \tilde{\delta}^{(\varepsilon,l)}(Z) - \sum_{i=1}^{N_g} \theta^{(\varepsilon,l)}_i \phi(Z;\kappa_i) \right)^\intercal f(Z) \right| \right] \\
& \le \left| \mathbb{E}_Z \left[ \left( \delta(Z) - \tilde{\delta}^\varepsilon(Z) \right) [f]_l (Z) \right] \right| + M \, \mathbb{E}_{Z} \left[ \left\| \delta^{(\varepsilon, l)}(Z) - \sum_{i=1}^{N_g} \theta^{(\varepsilon,l)}_i \phi(Z;\kappa_i)\right\|_2 \right] \\
& \le C_\delta \, \varepsilon \sup_{z\in\fR^k} \left(\big|[f]_l(z)\big| + \big\| \nabla [f]_l \big\| \right) + M \, \mathbb{E}_{Z} \left[ \left\| \tilde{\delta}^{(\varepsilon, l)}(Z) - \bv^{(\varepsilon,l)} - \int_{\fR^{n\times k}\times \fR^n} \phi (Z;\kappa) m_\xi^{(l)} (\kappa) \, d\kappa \right\|_2 \right] \\
& \quad + M\, \mathbb{E}_{Z} \left[ \left\| \bv^{(\varepsilon, l)} + \int_{\fR^{n\times k}\times \fR^n} \phi (Z;\kappa) m_\xi^{(l)} (\kappa) \, d\kappa - \sum_{i=1}^{N_g} \theta^{(\varepsilon,l)}_i \phi(Z;\kappa_i) \right\|_2 \right] \\
& \le \left( (C_\delta + 4)\varepsilon + \frac{\sqrt{n} \|\sigma_g\|_\infty C_\cK^{(l)}}{\sqrt{N_g-n}} \left( 1 + \sqrt{2 \log \frac{1}{\zeta}} \right) \right) \sup_{z\in\fR^k} \left(\|f(z)\|_2 + \|Df(z)\|\right)
\end{align*}
with probability $\ge 1-\zeta$.
Clearly, with sufficiently large $N_g$, the last term is $\cO(\varepsilon\,\sup_{z\in\fR^k} \left(\|f(z)\|_2 + \|Df(z)\|\right))$.

\section{Experimental details and additional experimental results}

\begin{figure*}[ht]
% \vspace{-5mm}
\centering
\begin{tabular}{ccc}
% \hspace{-7mm}
% \hspace{-9mm}
% \subfigure[Convergence of the loss function $L$]{
%       \includegraphics[width=0.45\textwidth]{fig2}}
% \end{tabular}\\
% \vspace{-7mm}
\hspace{-4mm}
\subfigure [Samples from true distribution $P_X$]{
      \raisebox{0.1 \height} {
      \includegraphics[width=0.3\textwidth]{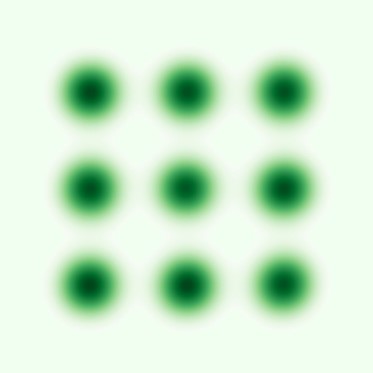}
      \label{subfig:9-true}
      }
      }
&
\hspace{-6mm}
\subfigure[Samples from generator $g_\theta(Z)$]{
      \raisebox{0.1 \height} {
      \includegraphics[width=0.3\textwidth]{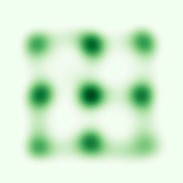}
      \label{subfig:9-gen}
      }}
&
\hspace{-6mm}
\subfigure[Convergence of the loss functions $J$ and $L$]{
      \raisebox{0.05 \height} {
      \includegraphics[width=0.38\textwidth]{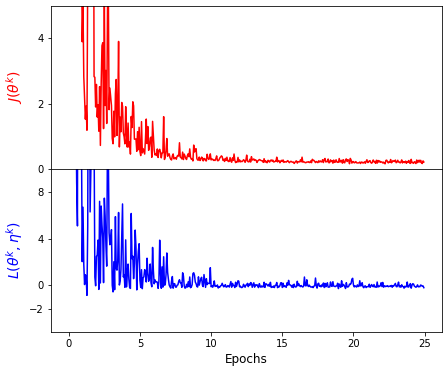}
      \label{subfig:9-loss}
      }
      }\\
\hspace{-4mm}
\subfigure [Samples from true distribution $P_X$]{
      \raisebox{0.1 \height} {
      \includegraphics[width=0.3\textwidth]{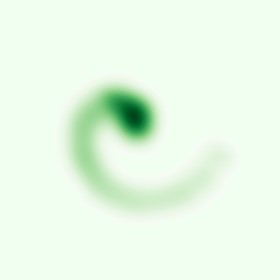}
      \label{subfig:spiral-true}
      }
      }
&
\hspace{-6mm}
\subfigure[Samples from generator $g_\theta(Z)$]{
      \raisebox{0.1 \height} {
      \includegraphics[width=0.3\textwidth]{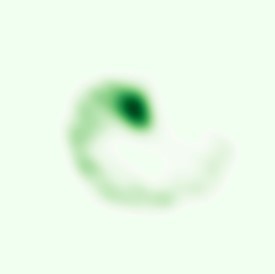}
      \label{subfig:spiral-gen}
      }}
&
\hspace{-6mm}
\subfigure[Convergence of the loss functions $J$ and $L$]{
      \raisebox{0.05 \height} {
      \includegraphics[width=0.38\textwidth]{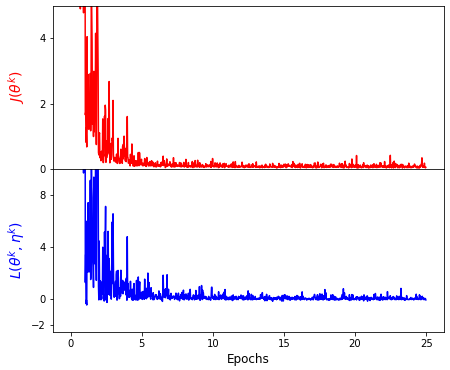}
      \label{subfig:spiral-loss}
      }
      }
\end{tabular}
 \vspace{-1.2em}
\caption{Additional experiments with mixtures of Gaussians. The code is available at \url{https://github.com/sehyunkwon/Infinite-WGAN}.} \vspace{-0.5em}
\label{fig:samples}
\end{figure*}

\subsection{Gaussian mixture sample generation}
We now provide details of the experiments for Figure~\ref{fig:samples}.
The code is available at \url{https://github.com/sehyunkwon/Infinite-WGAN}.
The true and latent distributions are 2-dimensional, i.e., $n=2$ and $k=2$.
The true distribution $P_X$ is a mixture of 8  Gaussians with equal weights, where the means are 
$(\sqrt{2}\cos \frac{m\pi}{4}, \sqrt{2}\sin \frac{m\pi}{4})$ for $m = 0, 1, \dots, 7$,
%$(\sqrt{2}, 0), (-\sqrt{2}, 0), (0, \sqrt{2}), (0, -\sqrt{2}), (1, 1), (1, -1), (-1, 1), (-1, -1)$
and the covariance matrices are $\begin{pmatrix}0.1^2& 0\\ 0 & 0.1^2\end{pmatrix}$.
The generator feature functions are of the form $\phi_i(x) =\sigma_g(\kappa_wz+\kappa_b)$ as described in \hyperlink{cond:AG}{(AG)}
for $1\leq i\leq N_g = 5000$, where the activation function $\sigma_g$ is $\tanh$.
Weights $\kappa_w$ and $\kappa_b$ for generator feature functions are randomly sampled (IID) from the Gaussian distribution
with zero mean and variance $10^2$.
As required in Lemma~\ref{lemma:finite-universal}, we create constant hidden units by replacing two sets of $(\kappa_w, \kappa_b)$ with $({\bf 0}_{2\times 2}, (1, 0))$ and $({\bf 0}_{2\times 2}, (0, 1))$.
The discriminator feature functions are of form $\psi_j(x) =\sigma(a_j^\intercal x + b_j)$ as described in \hyperlink{cond:AD}{(AD)}
for $1\leq j\leq N_d = 1000$ where the activation function $\sigma$ is $\tanh$.
We generate $a_j$ and $b_j$ independently according to the following procedure:
\begin{itemize}
\item Pick $x$-intercept $\tilde{a}$ and $y$-intercept $\tilde{b}$ from $-4$ to $4$ uniformly randomly.
\item Then, $\frac{x}{\tilde{a}} + \frac{y}{\tilde{b}} = 1$ is the line with those intercepts.
\item Pick $c$ uniformly randomly from $1$ to $10$, then set $a_j = (c/\tilde{a}, c/\tilde{b})$ and $b_j = -c$.
\end{itemize}
The generator stepsize starts at $\alpha=10^{-5}$ and decays by a factor of 0.9 at every epoch.
The networks are trained for 25 epochs with ($X$-sample) batch size 5000.
At each iteration, 5000 latent vectors ($Z$-samples) are sampled IID from the standard Gaussian distribution for the stochastic gradient ascent step,
and another 5000 latent vectors are sampled IID from the standard Gaussian distribution for the stochastic gradient descent step. 
The generator parameter $\theta$ is randomly initialized (IID) with the Gaussian distribution with zero mean and variance $5 \times 10^{-3}$.
The discriminator parameter $\gamma$ is randomly initialized (IID) with the standard normal distribution.
We visualize the generated distribution using the kernel density estimation (KDE) plot.

We also perform additional experiments under distinct settings.
The first additional experiment considers the true distribution $P_X$ that is a mixture of 9 Gaussians with equal weights. The means are
$(m_1, m_2)$ for $m_1=-1, 0, 1$ and $m_2=-1, 0, 1$, and the covariance matrices are the same as before.
We use the initial stepsize $\alpha=5 \times10^{-6}$ for the generator and $N_g=10,000$ for generator feature functions.
The generator parameter $\theta$ is randomly initialized (IID) with the Gaussian distribution with zero mean and variance $3 \times 10^{-3}$.
Discriminator feature functions are generated in the same manner.
Figure~\ref{subfig:9-true} shows the true distribution, and Figure~\ref{subfig:9-gen} shows the generated samples.
The second additional experiment considers the true distribution $P_X$, which is a spiral-shaped mixture of 20 Gaussians with equal weights.
The means are $(\frac{m}{20}\cos\frac{2m}{20}\pi, \frac{m}{20}\sin\frac{2m}{20}\pi)$ for $m=0, 1, \dots, 19$,
and the covariance matrices are the same as before.
We use the initial stepsize $\alpha=10^{-6}$ for the generator and $N_g=10,000$ for generator feature functions.
The generator parameter $\theta$ is randomly initialized (IID) with the Gaussian distribution with zero mean and variance $3 \times 10^{-3}$.
For the discriminator, feature function weights are generated by sampling $x$-intercept $\tilde{a}$ and $y$-intercept $\tilde{b}$ from $-2$ to $2$ uniformly randomly.
Figure~\ref{subfig:spiral-true} shows the true distribution, and Figure~\ref{subfig:spiral-gen} shows the generated samples.
In both cases, the generators closely mimic the true distributions and loss functions converge to zero.

\subsection{Loss landscape}
\begin{figure} [t!]
\centering
\begin{tabular}{cccc}
\includegraphics[width=0.34\textwidth]{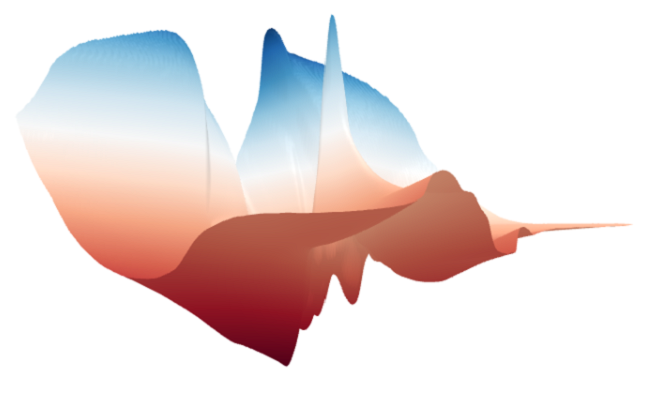} &
\includegraphics[width=0.29\textwidth]{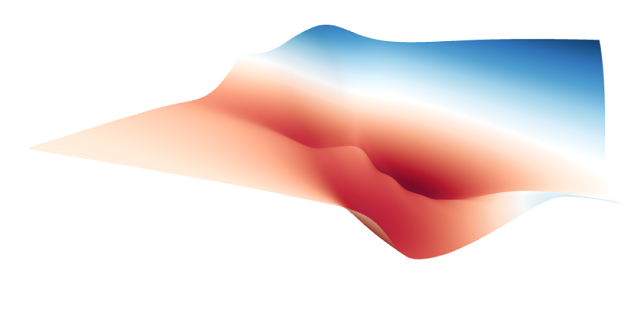} &
\includegraphics[width=0.29\textwidth]{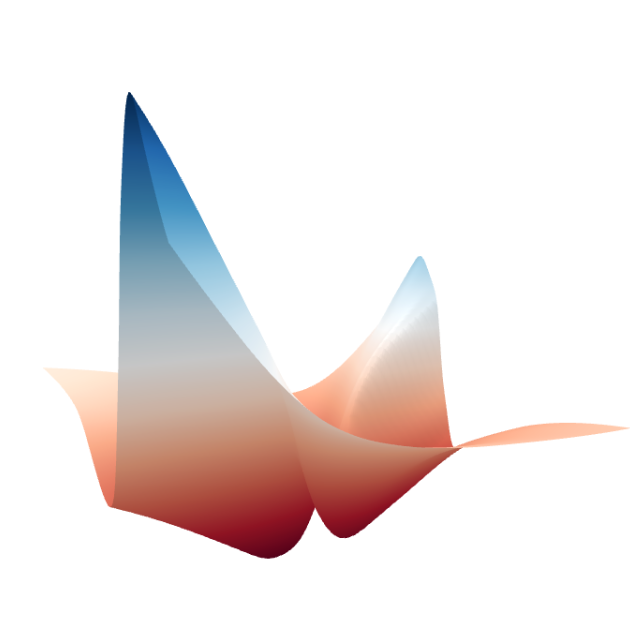} \\
\textbf{(a) $N_g=2$}  & \textbf{(b) $N_g=3$} & \textbf{(c) $N_g=5$}  \\[6pt]
\end{tabular}
\begin{tabular}{cccc}
\includegraphics[width=0.29\textwidth]{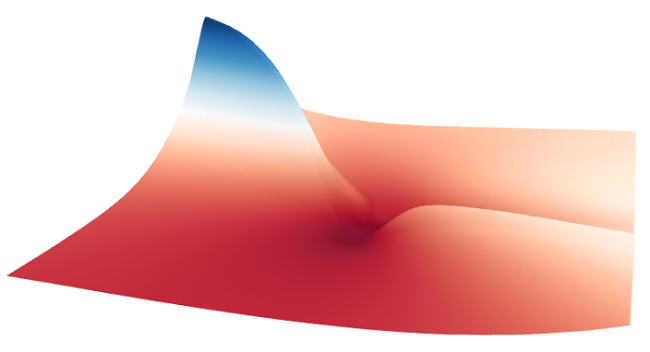} &
\includegraphics[width=0.39\textwidth]{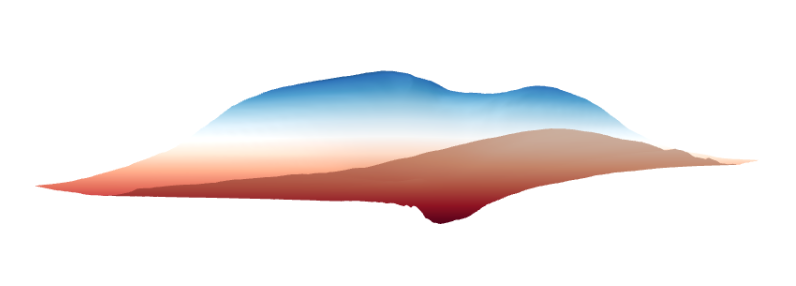} \\
\textbf{(d) $N_g=10$}  & \textbf{(e) $N_g=100$}  \\[6pt]
\end{tabular}
\caption{Loss landscapes of $J(\theta)$ for $N_g=2$, $3$, $5$, $10$, and $100$.}
\label{fig:landscape_appendix}
\vspace{0.5cm}
\centering
\begin{tabular}{cccc}
\includegraphics[width=0.29\textwidth]{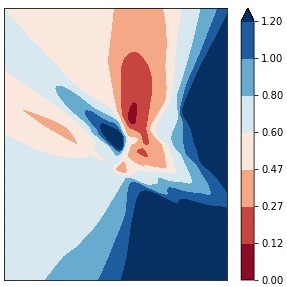} &
\includegraphics[width=0.29\textwidth]{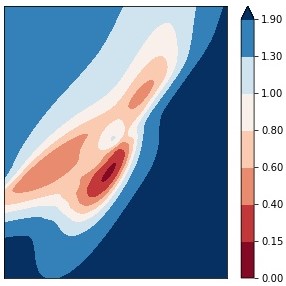} &
\includegraphics[width=0.29\textwidth]{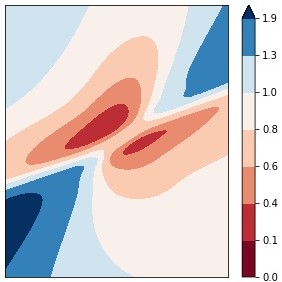} \\
\textbf{(a) $N_g=2$}  & \textbf{(b) $N_g=3$} & \textbf{(c) $N_g=5$}  \\[6pt]
\end{tabular}
\begin{tabular}{cccc}
\includegraphics[width=0.29\textwidth]{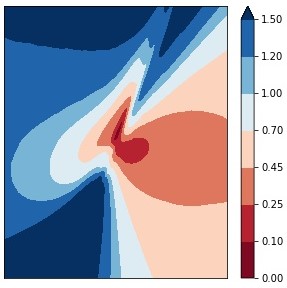} &
\includegraphics[width=0.29\textwidth]{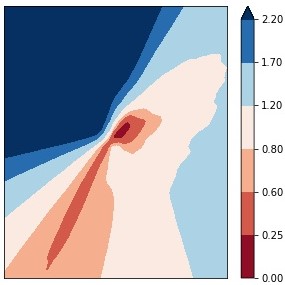} \\
\textbf{(d) $N_g=10$}  & \textbf{(e) $N_g=100$}  \\[6pt]
\end{tabular}
\vspace{-0.2cm}
\caption {Corresponding contour plots of the landscapes of Figure~\ref{fig:landscape_appendix}.}
\label{fig:contour_appendix}
\end{figure}

In this section, we describe the experiments for Figure~\ref{fig:landscape}, which visualizes the loss landscape of $J(\theta)$ for the cases $N_g=2$ and $N_g=10$.
We also provide additional experiments for $N_g=3$, $5$, and $100$. 
In the $N_g=2$ case, the landscape is highly non-convex and displays at least three non-global local minima.
We observe that in Figures~\ref{fig:landscape_appendix} and \ref{fig:contour_appendix}, the landscapes become better behaved, although still non-convex, as $N_g$ increases.

When $N_g>2$, the parameter space is projected down to a 2D plane spanned by two random directions, as recommended by \citet{li2018visualizing}. 
The true and latent distributions are 2-dimensional, i.e., $n=2$ and $k=2$. The true distribution $P_{X}$ is a mixture of 2 Gaussians with equal weights, where the means are $(m_1, m_2)$ for $m_1=0$ and $m_2 = \pm 2$, and the covariance matrices are $\begin{pmatrix} \sqrt{0.5^{2}} & 0 \\ 0 & \sqrt{0.5^{2}} \end{pmatrix}$. The latent distribution is the standard Gaussian distribution. The generator feature functions are of the form $\phi_i(x) =\sigma_g(\kappa_wz+\kappa_b)$ as described in \hyperlink{cond:AG}{(AG)} for $1\leq i\leq N_g = 2$, $3$, $5$, $10$, and $100$, where the activation function $\sigma_g$ is $\tanh$. Weights $\kappa_w$ are randomly sampled (IID) from an isotropic Gaussian and then multiplied by a scalar factor, sampled independently from the standard normal distribution. Weights $\kappa_b$ are randomly sampled (IID) from the Gaussian distribution with zero mean and variance $3 \times 10^{-1}$. The discriminator feature functions are of the form $\psi_j(x) =\sigma(a_j^\intercal x + b_j)$ as described in \hyperlink{cond:AD}{(AD)} for $1\leq j\leq N_d = 8$, where the activation function $\sigma$ is $\tanh$.

% Loss Landscape for Ng=2,3,5,10,100
\clearpage

\end{document}